\documentclass[a4paper,fleqn]{cas-dc}
\usepackage{subcaption}
\usepackage[numbers]{natbib}
\def\tsc#1{\csdef{#1}{\textsc{\lowercase{#1}}\xspace}}
\tsc{WGM}
\tsc{QE}
\tsc{EP}
\tsc{PMS}
\tsc{BEC}
\tsc{DE}
\begin{document}
\let\WriteBookmarks\relax
\def\floatpagepagefraction{1}
\def\textpagefraction{.001}

% Short title
\shorttitle{ }

% Short author
\shortauthors{Yanchen Guan et~al.}

% Main title of the paper
\title [mode = title]{Learning physically grounded traffic accident reconstruction from public accident reports}                      
% Title footnote mark
% eg: \tnotemark[1]

% Title footnote 1.
% eg: \tnotetext[1]{Title footnote text}
% \tnotetext[<tnote number>]{<tnote text>} 

% First author
%
% Options: Use if required
% eg: \author[1,3]{Author Name}[type=editor,
%       style=chinese,
%       auid=000,
%       bioid=1,
%       prefix=Sir,
%       orcid=0000-0000-0000-0000,
%       facebook=<facebook id>,
%       twitter=<twitter id>,
%       linkedin=<linkedin id>,
%       gplus=<gplus id>]
\author[1,2]{Yanchen Guan}\credit{Conceptualization of this study, Methodology, Experiment, Writing}
\author[1,3]{Haicheng Liao}\credit{Conceptualization of this study, Writing}
\author[1,2]{Chengyue Wang}\credit{Writing}
% \author[1]{Xingcheng Liu}\credit{Experiment}
% \author[1,2]{Jiaxun Zhang}\credit{Experiment}
% \author[4]{Keqiang Li}\credit{Conceptualization of this study}
\author[1,2,3]{Zhenning Li}[orcid=0000-0002-0877-6829]\cormark[1]\credit{Conceptualization of this study,Writing}\ead{zhenningli@um.edu.mo}

% % URL of the first author
% \ead[url]{www.cvr.cc, cvr@sayahna.org}

\cortext[cor1]{Corresponding author}
% \cortext[cor2]{Principal corresponding author}

% Address/affiliation
\affiliation[1]{organization={State Key Laboratory of Internet of Things for Smart City},
    addressline={University of Macau}, 
    city={Macau SAR},
    % citysep={}, % Uncomment if no comma needed between city and postcode
    postcode={999078}, 
    % state={},
    country={China}}
\affiliation[2]{organization={Department of Civil Engineering},
    addressline={University of Macau}, 
    city={Macau SAR},
    % citysep={}, % Uncomment if no comma needed between city and postcode
    postcode={999078}, 
    % state={},
    country={China}}
\affiliation[3]{organization={Department of Computer and Information Science},
    addressline={University of Macau}, 
    city={Macau SAR},
    % citysep={}, % Uncomment if no comma needed between city and postcode
    postcode={999078}, 
    % state={},
    country={China}}
% \affiliation[4]{organization={Department of Automotive Engineering},
%     addressline={Tsinghua University}, 
%     city={Beijing},
%     % citysep={}, % Uncomment if no comma needed between city and postcode
%     postcode={100084}, 
%     % state={},
%     country={China}}

% Here goes the abstract
\begin{abstract}
Traffic accidents are routinely documented in textual reports, yet physically grounded accident reconstruction remains difficult because detailed scene measurements and expert reconstructions are scarce, costly and hard to scale. Here we formulate accident reconstruction from publicly accessible reports and scene measurements as a parameterized multimodal learning problem. We construct CISS-REC, a dataset of 6,217 real-world accident cases curated from the NHTSA Crash Investigation Sampling System, and develop a reconstruction framework that grounds report semantics to road topology and participant attributes, reconstructs lane consistent pre-impact motion, and refines collision relevant interactions through localized geometric reasoning and temporal allocation. Our method outperforms representative baselines on CISS-REC, achieving the strongest overall reconstruction fidelity, including improved accident point accuracy and collision consistency. These results show that public accident reports can serve as scalable computational substrates for quantitatively verifiable accident reconstruction, with potential value for traffic safety analysis, simulation and autonomous driving research.
\end{abstract}
% We further generate controllable multi-view visualizations from the reconstructed trajectories.
% Use if graphical abstract is present
% \begin{graphicalabstract}
% \includegraphics{figs/grabs.pdf}
% \end{graphicalabstract}

% Research highlights
% \begin{highlights}
% \item Research highlights item 1
% \item Research highlights item 2
% \item Research highlights item 3
% \end{highlights}

% Keywords
% Each keyword is seperated by \sep

\maketitle

\section{Introduction}
\label{sec:Introduction}

Traffic crashes are among the most extensively documented events in modern transportation systems, yet they remain surprisingly difficult to study as physical processes~\cite{lord2010statistical,savolainen2011statistical}. In most large-scale accident archives, crashes are recorded as narratives: who was involved, what maneuver was attempted, where the collision occurred, and under what road and environmental conditions~\cite{arteaga2020injury}. But many of the questions that matter most to safety science are dynamic rather than narrative~\cite{lee2023advancing}. How did the vehicles move before impact? How did their interaction evolve? At what point did an initially feasible traffic situation become an inevitable collision? Answering such questions requires accidents to be represented not merely as reports, but as physically grounded motion histories.

This representational gap has become a major bottleneck for traffic safety research, forensic analysis, and autonomous-driving development. High-fidelity accident reconstruction can in principle recover scene-specific road geometry, pre-impact trajectories, speeds, and collision-relevant temporal interaction, but doing so usually requires reconstruction-grade evidence such as Event Data Recorder (EDR) logs, multi-view videos, detailed scene measurements, and expert post-crash analysis~\cite{tiwari2018transport}. Such evidence is expensive to collect, difficult to standardize, and often available only for a very small subset of crashes. Even when formal investigation systems exist, access to high-quality reconstructed cases is frequently limited by privacy, legal restrictions, heterogeneous investigation procedures, and the substantial manual effort required for scene interpretation~\cite{hu2021research,otte2012injury,kreiss2015extrapolation,montella2013crash}. As a result, physically measurable accident dynamics remain scarce precisely where scale is most needed.

By contrast, textual accident reports are produced routinely and at massive scale by police, investigators, insurers, and transportation agencies worldwide~\cite{zhang2018deep}. These reports are imperfect, but they are not uninformative. They often encode roadway context, participant maneuvers, travel directions, collision configuration, environmental conditions, and scene-level summaries that are directly relevant to crash formation. This creates a striking asymmetry in accident data: public archives are rich in semantic evidence but poor in dynamic, physically interpretable representations. If that asymmetry could be narrowed, accident reports could become more than administrative records. They could serve as computational resources for studying real-world crash dynamics at scale~\cite{shin2024recap}.

The difficulty, however, is fundamental rather than merely technical. A textual report does not directly specify trajectories, velocities, or precise timing, and the same narrative description may be compatible with many visually plausible scenes~\cite{beck2023automated}. In this sense, recovering accident dynamics from reports is an underdetermined inverse problem. Yet semantic plausibility is not enough. For accident data to support safety analysis, simulation, and autonomous-driving research, the recovered dynamics must be scene-specific, collision-consistent, and quantitatively assessable against measured evidence~\cite{wach2016calculation}. The challenge is therefore not simply to generate accident-like scenes from text, but to determine whether weak semantic descriptions can be transformed into physically meaningful and testable reconstructions of how crashes actually formed.

Recent progress in multimodal learning, simulation, and generative modeling has made this possibility increasingly compelling. Existing studies have shown that accident descriptions, sketches, or structured scene cues can support scene synthesis and visually plausible replay~\cite{guo2024sovar,tan2019text2scene,li2024steering}. These advances are important, but they do not by themselves resolve the central issue. Most current approaches remain closer to semantic restoration than to physically grounded motion recovery. They often rely on fixed templates, assumed initial conditions, approximate map priors, or perceptual realism objectives, and therefore do not recover scene-specific road topology, trajectory-level motion, and collision-consistent spatiotemporal dynamics in a quantitatively verifiable manner. What is still missing is a formulation that treats public accident reports not merely as descriptive narratives, but as weak yet scalable supervision for reconstructing physically meaningful pre-impact motion.

In this work, we argue that this problem becomes substantially more structured once report semantics are grounded to scene geometry and participant interaction. The key observation is that accident dynamics are not equally unconstrained over the entire pre-impact horizon. For much of the motion history, vehicle behavior is strongly regularized by roadway topology, lane assignment, travel direction, and maneuver intent. The decisive deviation that produces the crash is often concentrated near localized interaction between the collision-relevant participants. This view suggests that report-based accident reconstruction should not be framed as unrestricted scene generation. Instead, it can be posed as a constrained recovery problem in which global motion is shaped by road structure and behavioral intent, while crash formation is governed by localized geometric and temporal interaction.

Based on this view, we formulate accident reconstruction from public accident reports as a weakly supervised multimodal inverse problem. Rather than treating reports as purely narrative descriptions, we treat them as scalable supervisory carriers whose information can be progressively grounded through road topology, participant attributes, and collision structure to recover pre-impact vehicle motion. To support this problem, we construct CISS-REC, a real-world benchmark derived from the National Highway Traffic Safety Administration (NHTSA) Crash Investigation Sampling System (CISS), which systematically organizes public accident reports, scene measurements, road geometry, and motion-related evidence into a unified reconstruction dataset~\cite{griffin2020automatic}. Building on this benchmark, we develop a scene-grounded reconstruction framework that jointly models report semantics, road structure, participant attributes, and collision-localized interaction to reconstruct lane-consistent pre-impact motion and refine accident-critical geometric deviations under temporal and physical constraints.

Our contributions are threefold. First, we introduce a new formulation of accident reconstruction in which public accident reports are treated as weak but scalable supervision for physically grounded motion recovery, shifting the focus from semantic replay toward quantitatively verifiable reconstruction. Second, we construct CISS-REC, a large-scale real-world benchmark that enables systematic study of scene-specific accident reconstruction from heterogeneous semantic and geometric evidence. Third, we propose a scene-grounded reconstruction framework that exploits the structural decomposition of accident dynamics into globally constrained motion and localized collision-critical interaction, and we show that this formulation improves reconstruction fidelity over representative baselines.

More broadly, this work suggests a different role for public accident archives in transportation research. Rather than viewing them only as textual records of completed events, we show that they can be used as scalable substrates for recovering physically interpretable crash dynamics. This opens a path toward studying accident mechanisms, generating reconstruction-grade safety data, and supporting downstream applications in simulation, safety analysis, and autonomous driving.

\section{Related work}
\label{sec:Related}

\noindent\textbf{Forensic accident reconstruction and its scaling limits.}
Traffic accident reconstruction has traditionally been studied as a forensic and engineering problem aimed at determining how a crash occurred and which factors contributed to its outcome~\cite{zheng2020determinants,struble2020automotive,fernandes2018application,ryan2024accident}. In conventional practice, investigators reconstruct accident processes by combining scene measurements, photography, sketches, vehicle inspections, witness interviews, and in-vehicle data retrieval into a structured account of pre-impact motion and collision formation~\cite{rivers2010technical,rivers2006evidence}. Standard workflows typically involve evidence preservation, roadway measurement, coordinate establishment, geometric reconstruction, and expert post-analysis~\cite{mohammed2023overview,vida2023analysis,su2016developing,jiang2021unmanned,lemmens2011terrestrial,scherer2009conventional,clamann2021advancing,dhanam2025event}. These procedures can yield physically meaningful and case-specific reconstructions, but they are difficult to scale because they require specialized expertise, dedicated equipment, and substantial on-site effort. As a result, high-quality reconstruction-grade accident data remain scarce relative to the much larger population of real-world crashes~\cite{carper2000forensic,raviv2017analyzing,faizan2021forensic,smith1957physical}.

\noindent\textbf{The gap between large-scale accident archives and reconstruction-grade data.}
This scarcity has become a central bottleneck for data-driven accident analysis~\cite{liu2023integrated}. Detailed reconstruction is impractical for the vast majority of crashes, especially when accurate measurement, manual interpretation, and expert validation are required. In addition, reconstruction data often involve sensitive legal and personal information, which further limits accessibility and standardization~\cite{gadotti2024anonymization,morehouse2024responsible}. Even official in-depth investigation systems release only a small fraction of crashes in reconstruction-ready form, and the resulting records may still differ in completeness, consistency, or reliability~\cite{hossain2023data}. Taken together, these factors have produced a persistent mismatch between the scale at which crashes are documented and the scale at which physically interpretable accident dynamics are available for computational study.

\noindent\textbf{From narrative evidence to structured accident scenes.}
Because exhaustive reconstruction is infeasible in routine practice, most crashes are documented primarily through textual reports~\cite{font2012reconstruccion,rider2017impact,ball2005working,komter2006talk}. These reports were not designed for motion recovery, yet they encode information that is highly relevant to accident formation, including roadway context, participant maneuvers, travel directions, collision configuration, and environmental conditions. This has motivated a recent line of research that attempts to convert textual or weakly structured accident evidence into structured scene representations~\cite{chen2025transforming,jiao2018virtual}.

One direction combines textual reports with simulation priors. SoVAR~\cite{guo2024sovar}, for example, uses accident reports together with predefined road-structure maps to generate candidate trajectories and simulate the resulting scenes. Such approaches demonstrate that textual reports can support scene generation, but they typically depend on fixed road templates and assumption-driven initial conditions rather than on scene-specific physical evidence. The generated scenes may therefore satisfy coarse semantic consistency without constituting faithful reconstruction of individual real-world accidents.

A second direction emphasizes perceptual realism through generative modeling. AVD2~\cite{li2025avd2} reconstructs accident processes from semantic descriptions using diffusion-based video generation, while AccidentSim~\cite{zhang2025accidentsim} leverages large language models to translate report narratives into simulation cues and accident videos. These methods are valuable for producing visually plausible scenes and semantically coherent replay, but their outputs are primarily perceptual rather than reconstruction-grade. They do not explicitly enforce strict correspondence to real road geometry, measured motion, or collision-consistent spatiotemporal dynamics, which makes it difficult to regard them as physically faithful reconstruction of specific accidents.

A third direction uses sketches or other intermediate visual representations as stronger structural cues. CrashAgent~\cite{li2025crashagent}, for instance, employs vision--language models to generate simulator-ready scenes from accident sketches, including sketched trajectories and keypoints. This formulation is appealing because sketches often contain stronger spatial information than free-form narratives. However, it relies on the availability of complete and informative sketches, which are uncommon in many public accident archives. Moreover, road layouts are typically recovered by matching recognized road categories to parameterized templates, so the resulting topology aligns with the true scene only at a coarse categorical level rather than through strict geometric grounding.

\noindent\textbf{What is still missing.}
Taken together, existing studies show that public accident records can support scene synthesis, semantic replay, and visually plausible accident simulation. However, most current approaches still stop short of physically grounded and quantitatively verifiable recovery of crash dynamics. They generally rely on fixed templates, assumed initial conditions, approximate map priors, or perceptual realism objectives, and therefore remain limited in scene specificity, geometric grounding, and physical testability. In other words, they help bridge the gap from narrative evidence to plausible scene generation, but not yet the more difficult gap from narrative evidence to reconstruction-grade dynamic representation. The present work is motivated precisely by this missing step: how to recover scene-specific pre-impact motion from public accident archives in a manner that is not only semantically consistent, but also geometrically grounded, collision-consistent, and assessable against measured evidence.

\section{Methodology}

\subsection{Problem setup}
\label{setup}
We formulate accident reconstruction as the recovery of scene specific vehicle motion from sparse report semantics and measured road geometry. For each accident case, the model is conditioned on two types of input evidence: report-derived semantic information and scene geometry. The goal is to infer a physically plausible and semantically consistent pre-impact scene whose motion evolution remains compatible with the accident description, road topology, and collision context.

Formally, let the model input for one accident case be denoted by
$\mathcal{X} = (\mathcal{R}, \mathcal{G})$,
where $\mathcal{R}$ denotes report-derived semantic information, including the global crash summary and vehicle-level attributes, and $\mathcal{G}$ denotes scene geometry, including road curves, lane centerlines, and lane polygons. Suppose that the case involves $N$ vehicles and that the pre-impact horizon is discretized into $K$ time steps. For vehicle $i \in \{1,\dots,N\}$, the target pre-impact trajectory is represented as
$\mathcal{T}_i = \{(\mathbf{p}_{i,k}, v_{i,k})\}_{k=1}^{K}$,
where $\mathbf{p}_{i,k} \in \mathbb{R}^{2}$ denotes the planar position of vehicle $i$ at time step $k$, and $v_{i,k}$ denotes the corresponding speed.

The reconstruction model aims to reconstruct
$\hat{\mathcal{T}} = \{\hat{\mathcal{T}}_i\}_{i=1}^{N}$,
such that the reconstructed trajectories remain close to the reference motion under the available supervision, while preserving consistency with behavior-related cues, collision relevant interaction, and scene-specific road geometry.

During training, the model is supervised using sparse motion annotations,
$\mathcal{S} = (\mathcal{P}^{\mathrm{obs}}, \mathcal{A}, \mathcal{V}^{\mathrm{edr}})$,
where $\mathcal{P}^{\mathrm{obs}}$ denotes ordered pre-impact investigation points, $\mathcal{A}$ denotes impact-related annotations, and $\mathcal{V}^{\mathrm{edr}}$ denotes EDR-derived speed cues. Among these signals, $\mathcal{P}^{\mathrm{obs}}$ and $\mathcal{V}^{\mathrm{edr}}$ are used only to construct reconstruction supervision during training, rather than as model inputs at inference time. Under this formulation, accident reconstruction is cast as a parameterized multimodal learning problem that maps sparse semantic and geometric evidence to quantitatively verifiable pre-impact motion.

\subsection{Input representation}

\begin{figure}[t]
\centering 
\includegraphics[width=0.48\textwidth]{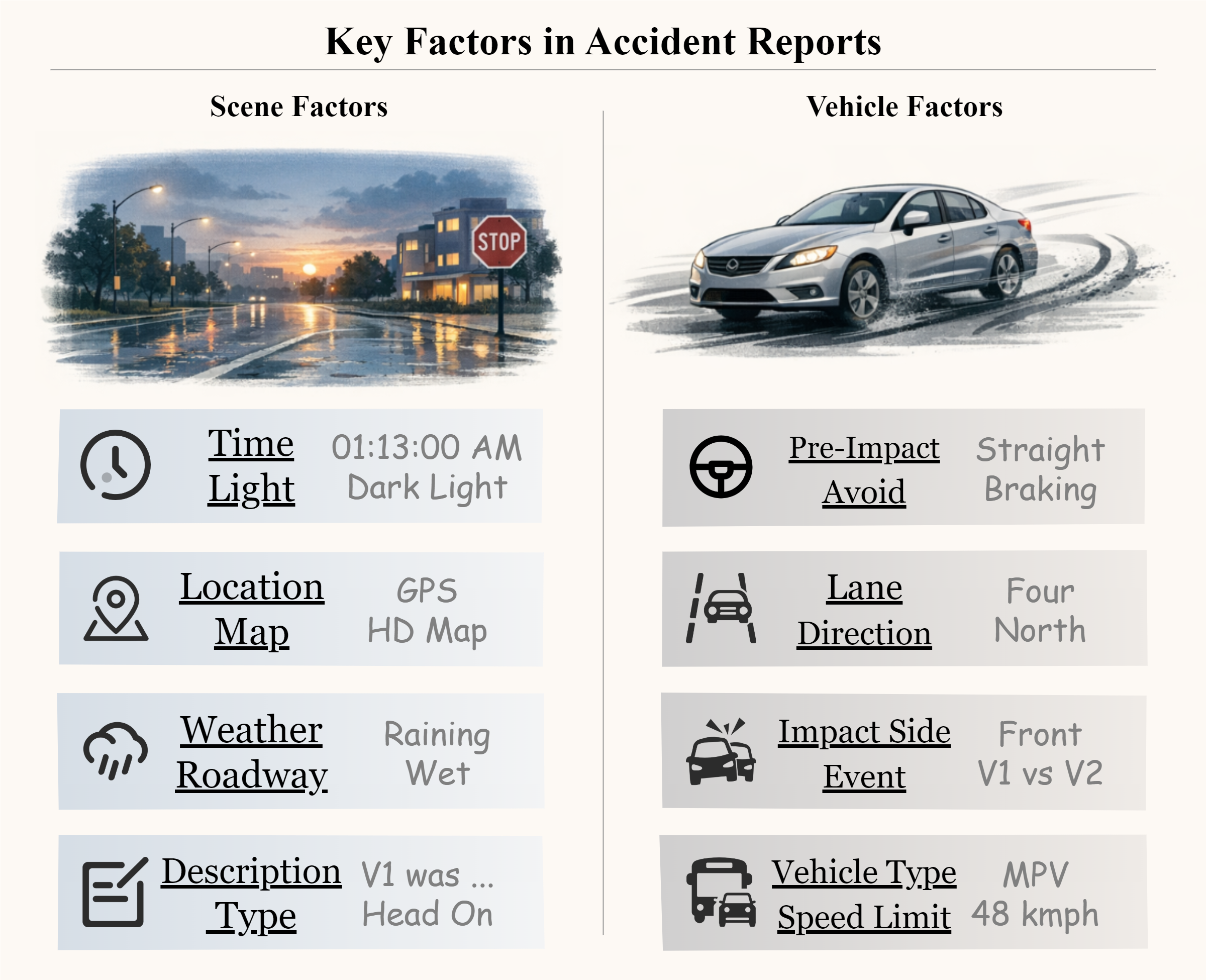}
\caption{The input representations and corresponding input examples derived from the original accident report. In the same traffic accident report, scene factors are global attributes, while vehicle factors are individual attributes of the accident participants.}
\label{input}
\end{figure}

To reconstruct motion from accident reports, we organize the model input into three complementary components: scene-level semantics, vehicle-level semantics, and scene geometry. This decomposition follows the native structure of accident reports, in which global scene descriptions and vehicle-specific records are provided separately but jointly constrain accident evolution.

The scene-level semantic input summarizes the global accident context shared by all involved vehicles. It includes the crash summary together with scene attributes that are relevant to reconstruction, such as accident time, lighting condition, weather, roadway condition, and geographic context. These variables provide coarse but globally informative constraints on the reconstructed scene.

The vehicle-level semantic input describes the motion-related attributes of each involved vehicle. For vehicle $i$, these attributes include its initial travel lane, travel direction, pre-impact movement, attempted avoidance maneuver, vehicle category, and posted speed limit. Collectively, these variables specify the local behavioral cues that condition the reconstruction of vehicle-specific motion.

The geometric input describes the scene-specific road structure. It consists of standardized road curves together with lane centerlines and lane polygons, which provide explicit spatial support for grounding report semantics to the physical scene. In particular, the combination of lane assignment, travel direction, and road geometry constrains the feasible spatial initialization and subsequent motion of each vehicle.

Figure~\ref{input} illustrates the key factors selected as input for reconstruction of traffic accidents at the scene-level and vehicle-level. Under this representation, an accident case is encoded as
$\mathcal{X} = (\mathcal{R}^{\mathrm{scene}}, \mathcal{R}^{\mathrm{veh}}, \mathcal{G})$,
where $\mathcal{R}^{\mathrm{scene}}$ denotes scene-level semantic descriptors, $\mathcal{R}^{\mathrm{veh}} = \{\mathcal{R}^{\mathrm{veh}}_i\}_{i=1}^{N}$ denotes vehicle-level semantic descriptors for the $N$ involved vehicles, and $\mathcal{G}$ denotes scene geometry. This representation preserves the semantic hierarchy of the original report while exposing the global context, vehicle-specific cues, and spatial structure required for scene-grounded accident reconstruction.

\subsection{Overall framework}

The proposed framework reconstructs accident scenes from report semantics and scene geometry in three stages: input standardization, scene-grounded motion reconstruction, and downstream visual rendering. First, heterogeneous accident evidence is converted into a unified representation that aligns semantic descriptions, road structure, and motion-related annotations within a common spatial and temporal frame. This step provides the structured input and supervision required for learning-based reconstruction.

Second, based on the standardized representation, we develop an encoder--decoder architecture for scene-grounded accident reconstruction. The encoder integrates scene-level semantics, vehicle-level semantic descriptors, and scene geometry into a shared latent representation that associates report semantics with road structure and participating vehicles. Conditioned on this representation, the decoder reconstructs motion for all involved vehicles, and further refines collision relevant interaction through localized geometric reasoning and temporally constrained motion allocation. As a result, the model jointly captures global scene context, vehicle-specific behavioral cues, and accident-critical interaction patterns within a unified reconstruction process.

Finally, the reconstructed trajectories can be coupled with a downstream rendering pipeline to produce controllable visualizations of the accident scene from multiple viewpoints. In this way, the framework links report-conditioned motion reconstruction to visually interpretable accident replay while preserving consistency with the reconstructed scene dynamics.

\subsection{Data preprocessing}
\label{preprocessingtex}

\begin{figure*}[t]
\centering 
\includegraphics[width=0.70\textwidth]{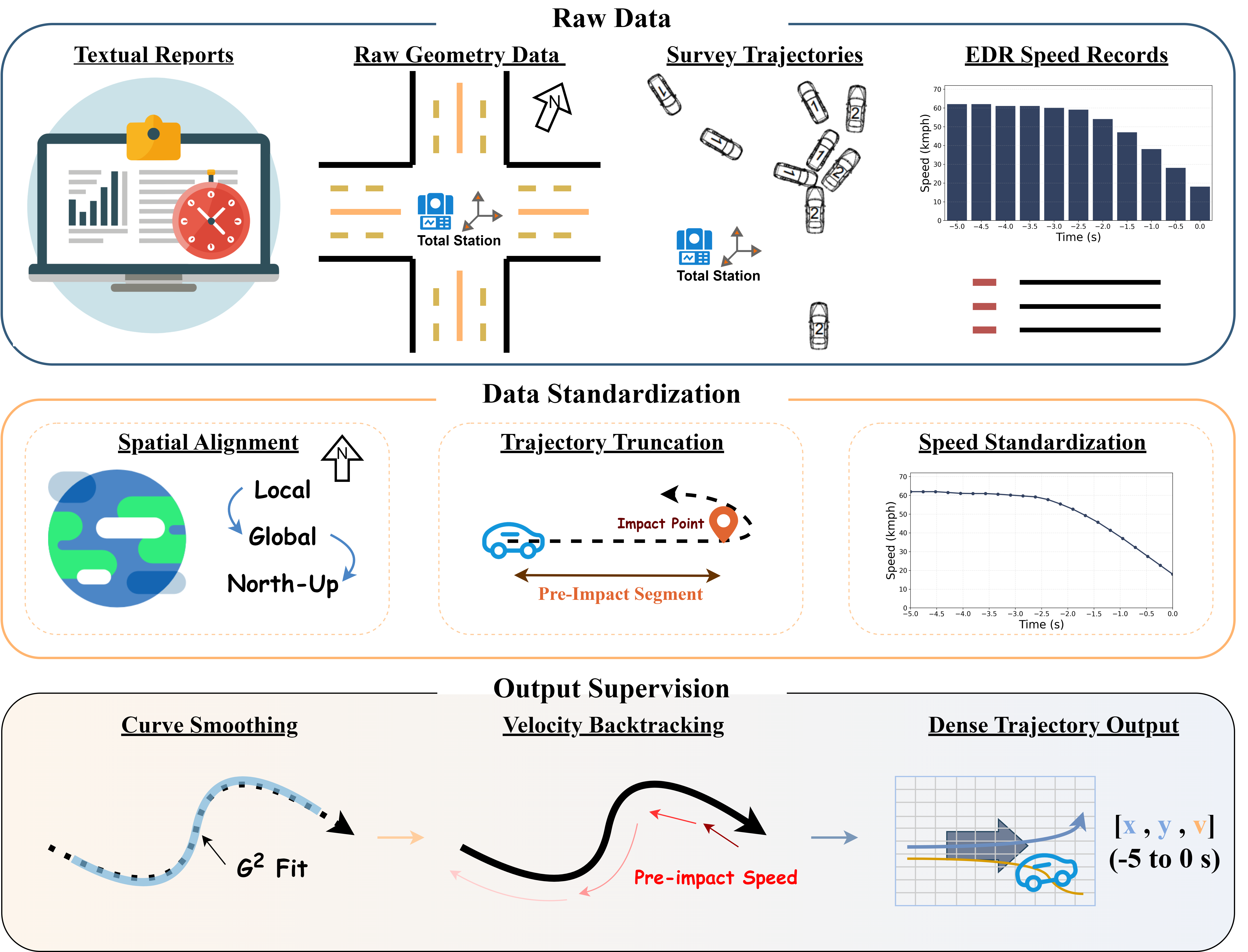}
\caption{The preprocessing of velocity and trajectory data. Sparse trajectories and EDR records in the original accident reports are transformed into temporally dense supervision through alignment, fitting, and reconstruction, providing structured training targets for accident reconstruction.}
\label{preprocessing}
\end{figure*}

We convert each accident case into a standardized representation by jointly processing report semantics, scene geometry, and motion annotations. The purpose of preprocessing is twofold: to align heterogeneous accident evidence within a common spatial frame, and to construct dense reference supervision from sparse measurements. Under this design, semantic inputs used for reconstruction are kept separate from motion annotations used only during training.

We first standardize the semantic and geometric content of each case. From the original accident reports, we retain only the variables that directly constrain accident reconstruction, including the global crash summary and vehicle-level attributes such as initial travel lane, travel direction, posted speed limit, pre-impact movement, and attempted avoidance maneuver, while normalizing missing entries as unknown to maintain a consistent schema across cases. Raw measurement geometry is then transformed from the local total station coordinate system to a common world coordinate system using the recorded rigid transformation. All map and trajectory entities are further rotated into a north-up convention, with the rotation angle determined preferentially from north-arrow annotations and otherwise from default scene metadata. This step aligns road curves, lane geometry, fitted keypoints, and trajectory evidence across accident cases under a unified spatial convention.

To support scene-grounded reconstruction, road geometry is represented as fixed-size tensors of ordered polylines with point-wise semantic category labels, while lane centerlines and lane polygons are extracted in parallel to provide lane-level structural support. Vehicle records are aligned with case-level summaries to form a fixed number of participant slots with harmonized semantic attributes. After this step, each accident case is represented by standardized scene-level semantics, vehicle-level semantic descriptors, and scene geometry, which together constitute the model input defined in Section~\ref{setup}.

Because the original accident reports provide sparse and heterogeneous motion evidence rather than dense trajectories, we further construct dense reference supervision for training by combining impact-related annotations, ordered investigation trajectories, and EDR-derived speed measurements when available. As shown in Figure~\ref{preprocessing}, sparse EDR observations are interpolated onto a uniform temporal grid covering the 5~s interval before impact, yielding a normalized speed sequence at 0.1~s resolution. The availability and temporal coverage of valid EDR records are retained as reliability indicators for supervision.

For each vehicle, ordered investigation points are aligned with the annotated impact location and truncated at the collision related endpoint so that only the physically relevant pre-impact segment is retained. When an exact impact-location match is unavailable, the last valid observation point is used as a fallback endpoint to preserve supervision continuity. The resulting sparse path is then regularized by piecewise \(G^2\)-continuous curve fitting, yielding a smooth reference path with continuous heading and curvature~\cite{bertolazzi2018g2}.

Formally, let the interpolated speed sequence of vehicle \(i\) be denoted by
\begin{equation}
\mathcal{V}_i = \{(t_k, v_{i,k})\}_{k=1}^{K}, \qquad t_k \in [-5,0],
\end{equation}
where the timestamps are uniformly sampled at 0.1~s intervals. Let
\begin{equation}
\mathcal{P}_i = \{(x_{i,n}, y_{i,n}, \theta_{i,n})\}_{n=1}^{N_i},
\end{equation}
denote the ordered pre-impact trajectory points after impact-aligned truncation, and let \(\tilde{\mathcal{P}}_i(\ell)\) denote the corresponding \(G^2\)-continuous reference curve parameterized by arc length \(\ell\). Taking the impact location as the spatial anchor at \(t=0\), the cumulative backward travel distance at time \(t_k\) is computed as
\begin{equation}
\ell_{i,k} = \int_{t_k}^{0} \hat{v}_i(t)\, dt,
\end{equation}
where \(\hat{v}_i(t)\) denotes the interpolated speed profile. The dense pre-impact reference supervision is then written as
\begin{equation}
\mathcal{T}_i = \{(t_k, v_{i,k}, x_{i,k}, y_{i,k}, \theta_{i,k})\}_{k=1}^{K},
\end{equation}
where \((x_{i,k}, y_{i,k}, \theta_{i,k})\) is obtained by tracing backward along \(\tilde{\mathcal{P}}_i\) according to \(\ell_{i,k}\).

For vehicles without sufficiently reliable EDR measurements, we derive weak speed-conditioned supervision from the reported speed limit and retain both vehicle-level and timestep-level validity masks to distinguish strongly constrained supervision from weakly constrained estimates. Additional accident-related annotations, including impact-side labels, collision pair cues, and other reliability-aware masks, are used only during training to support supervision construction and auxiliary objectives. After preprocessing, each accident case therefore provides standardized semantic and geometric inputs for reconstruction, together with dense and reliability-aware reference supervision for model training.

\subsection{Semantic scene encoder}

\begin{figure*}[t]
\centering 
\includegraphics[width=0.98\textwidth]{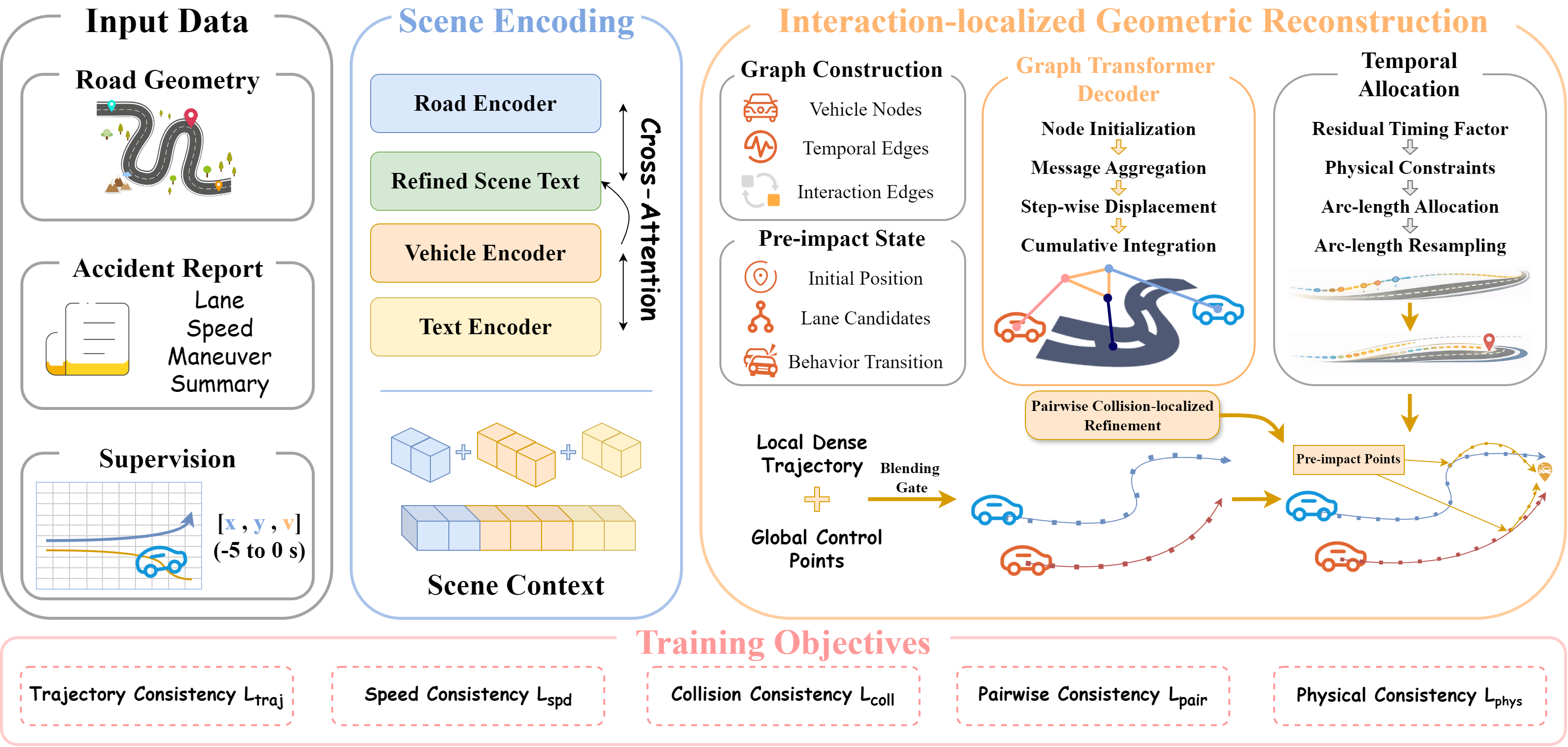}
\caption{Overall framework of our physically grounded accident reconstruction model.
The overall architecture of the proposed model follows an encoder--decoder paradigm. The input scene is first encoded into a latent representation space. Based on this representation, a graph structure is constructed to model interactions among multiple traffic participants. The decoder then generates candidate trajectories from the latent space, which are subsequently refined through accident-specific constraints and temporal reordering.}
\label{framework}
\end{figure*}

Following preprocessing, each accident case is represented by standardized scene-level semantics, vehicle-level semantic descriptors, and scene geometry as shown in Figure~\ref{framework}. Because accident reports do not directly provide dense trajectories, the encoder maps heterogeneous semantic and geometric inputs into a shared latent scene representation from which reconstruction can be inferred. The key objective of the encoder is therefore to ground report semantics to both participating vehicles and the surrounding road structure.

Road geometry is encoded through two complementary streams. First, the standardized road curves are rasterized into a multi-channel bird's-eye-view (BEV) tensor, in which different semantic categories are assigned to separate channels. A convolutional backbone extracts dense road tokens from this raster representation, capturing local occupancy patterns, geometric continuity, and nearby spatial support. Second, each ordered road polyline is summarized into a compact geometric descriptor containing its midpoint, dominant orientation, normalized length, spatial extent, and semantic class, and is then projected into a sparse shape token. The final road-token sequence is obtained by concatenating the dense BEV tokens and sparse shape tokens, thereby preserving both local geometric detail and higher-level road organization. A pooled road representation is further obtained by masked aggregation over the full road-token sequence:
\begin{equation}
\mathbf{r}^{\mathrm{global}} = \mathrm{Pool}\!\left(\{\mathbf{r}_n\}\right).
\end{equation}

Each vehicle is encoded from its textual and structured descriptors. Let $\mathbf{e}^{\mathrm{veh}}_i$ denote the vehicle-level text embedding, and let $\mathbf{b}_i$ denote its structured semantic attributes, including initial travel lane, travel direction, pre-impact movement, attempted avoidance maneuver, vehicle category, and posted speed limit. These features are fused into a vehicle token,
\begin{equation}
\mathbf{z}_i = \Psi_{\mathrm{veh}}(\mathbf{e}^{\mathrm{veh}}_i, \mathbf{b}_i).
\end{equation}
Only valid participant slots contribute to downstream aggregation. Masked pooling over valid vehicles yields a global participant representation,
\begin{equation}
\mathbf{z}^{\mathrm{global}} =
\frac{1}{|\mathcal{I}_{\mathrm{valid}}|}
\sum_{i \in \mathcal{I}_{\mathrm{valid}}} \mathbf{z}_i.
\end{equation}

The global crash summary is encoded in parallel as a scene-level textual descriptor $\mathbf{t}$. To align report semantics with scene entities, the encoder applies two successive cross-attention operations. The scene text first attends to the vehicle-token sequence, allowing the global narrative to focus on participant-specific evidence, and then attends to the road-token sequence, aligning the refined semantics with scene geometry:
\begin{equation}
\begin{split}
\tilde{\mathbf{t}}^{\mathrm{veh}}
=
\mathbf{t}
+
\mathrm{MHA}\!\left(
\mathbf{t},
\{\mathbf{z}_i\},
\{\mathbf{z}_i\}
\right),\\
\tilde{\mathbf{t}}^{\mathrm{road}}
=
\tilde{\mathbf{t}}^{\mathrm{veh}}
+
\mathrm{MHA}\!\left(
\tilde{\mathbf{t}}^{\mathrm{veh}},
\{\mathbf{r}_n\},
\{\mathbf{r}_n\}
\right).
\end{split}
\end{equation}

This hierarchical grounding reflects the structure of accident reports, in which the global narrative primarily describes vehicle-level behavior and is subsequently interpreted within the surrounding road environment.

Finally, the pooled road representation, pooled participant representation, refined scene text, and scene-level speed-related cues are fused to form the scene context:
\begin{equation}
\mathbf{c}
=
\Psi_{\mathrm{fuse}}
\!\left(
\mathbf{r}^{\mathrm{global}},
\mathbf{z}^{\mathrm{global}},
\tilde{\mathbf{t}}^{\mathrm{road}},
\mathbf{u}^{\mathrm{spd}}
\right),
\end{equation}
where $\mathbf{u}^{\mathrm{spd}}$ denotes the encoded speed-related cues. The encoder therefore outputs a scene context vector $\mathbf{c}$, a sequence of vehicle tokens $\{\mathbf{z}_i\}$, and a sequence of road tokens $\{\mathbf{r}_n\}$, together with the corresponding validity masks. These latent variables provide the common conditioning interface for following geometric reconstruction and temporal allocation.

\subsection{Interaction-localized geometric reconstruction}

Conditioned on the encoded scene representation, the decoder reconstructs pre-impact motion in two stages: a globally lane consistent trajectory reconstruction stage and a localized collision refinement stage. This design reflects the structure of accident formation. For most of the pre-impact horizon, vehicle motion is primarily constrained by lane geometry and report-derived intent, whereas the decisive geometric deviation typically occurs in the vicinity of collision relevant interaction.

For each vehicle \(i\), the decoder first predicts a coarse start location from its vehicle token, the scene context, and the corresponding structured semantic descriptors:
\begin{equation}
\tilde{\mathbf{p}}_{i,0}
=
\Psi_{\mathrm{start}}(\mathbf{z}_i,\mathbf{c},\mathbf{b}_i).
\end{equation}

Because the coarse prediction may not yet be spatially aligned with the road layout, it is further grounded to scene-specific lane priors. Let \(\mathcal{L}_m\) denote the \(m\)-th candidate lane centerline, and let \(\mathbf{u}_i\) and \(\mathbf{u}_m\) denote the reported travel direction of vehicle \(i\) and the local tangent direction of lane \(\mathcal{L}_m\), respectively. Candidate lanes are scored jointly by spatial proximity and directional consistency:
\begin{equation}
\mathrm{score}(i,m)
=
-
\Bigl[
d\!\left(\tilde{\mathbf{p}}_{i,0},\mathcal{L}_m\right)
-
\lambda_{\mathrm{dir}}
\max\!\bigl(0,\cos \angle(\mathbf{u}_i,\mathbf{u}_m)\bigr)
\Bigr].
\end{equation}

The top-ranked candidates are converted into forward anchor paths,
\begin{equation}
\mathbf{A}^{(m)}_i = \{\mathbf{a}^{(m)}_{i,k}\}_{k=1}^{K},
\end{equation}
and a learned mixture model assigns a probability to each candidate hypothesis:
\begin{equation}
\pi^{(m)}_i
=
\frac{\exp(\eta^{(m)}_i)}
{\sum_{m'} \exp(\eta^{(m')}_i)}.
\end{equation}

The selected lane prior is then used to project the coarse start state onto the lane consistent initialization \(\mathbf{p}_{i,0}\), which serves as the geometric origin of downstream reconstruction.

Based on the aligned start state and lane anchor cues, the decoder reconstructs a globally lane consistent base trajectory. A spacetime graph is built over valid vehicles and time steps, where each node corresponds to one vehicle at one time step. Node representations combine the vehicle token with anchor-derived motion cues, while temporal edges connect adjacent steps of the same vehicle and interaction edges connect nearby vehicles. A graph transformer (GT) propagates information on this sparse spacetime graph:
\begin{equation}
\mathbf{h}^{(\ell+1)}_{i,k}
=
\mathrm{GT}^{(\ell)}
\!\left(
\mathbf{h}^{(\ell)}_{i,k},
\mathcal{N}(i,k),
\mathbf{e}_{(i,k)\rightarrow(j,k')}
\right).
\end{equation}

The updated node states are decoded into step-wise displacement increments,
\begin{equation}
\Delta \mathbf{p}_{i,k}
=
\Psi_{\mathrm{disp}}(\mathbf{h}^{(L)}_{i,k}),
\end{equation}
whose cumulative integration yields a dense motion branch,
\begin{equation}
\mathbf{p}^{\mathrm{dense}}_{i,k}
=
\mathbf{p}_{i,0}
+
\sum_{\kappa=1}^{k}
\Delta \mathbf{p}_{i,\kappa}.
\end{equation}

To improve global path regularity, the decoder additionally predicts a sparse set of control points,
\begin{equation}
\{\mathbf{q}_{i,r}\}_{r=1}^{R}
=
\Psi_{\mathrm{ctrl}}(\mathbf{z}_i,\mathbf{c}),
\end{equation}
which are interpolated into a smooth fitted path,
\begin{equation}
\mathbf{p}^{\mathrm{fit}}_{i,k}
=
\mathrm{Interp}\!\left(\{\mathbf{q}_{i,r}\}_{r=1}^{R}\right).
\end{equation}

A learned blending gate then combines the locally flexible dense branch with the globally regular fitted branch:
\begin{equation}
\mathbf{p}^{\mathrm{base}}_{i,k}
=
(1-\gamma_i)\mathbf{p}^{\mathrm{dense}}_{i,k}
+
\gamma_i \mathbf{p}^{\mathrm{fit}}_{i,k}.
\end{equation}

In parallel, the decoder predicts a transition variable \(\hat{\tau}_i\) for each vehicle,
\begin{equation}
\hat{\tau}_i
=
\Psi_{\tau}(\mathbf{z}_i,\mathbf{c},\mathbf{b}_i),
\end{equation}
which represents the onset of accident-relevant motion change.

Because the geometric deviation that determines collision formation is typically concentrated on the collision relevant pair, the decoder further applies a localized interaction-refinement branch. Let \((i,j)\) denote the target collision pair derived from accident annotations. The base relative motion of this pair is written as
\begin{equation}
\mathbf{r}^{\mathrm{base}}_{ij,k}
=
\mathbf{p}^{\mathrm{base}}_{j,k}
-
\mathbf{p}^{\mathrm{base}}_{i,k}.
\end{equation}

A pair token is then constructed from the two vehicle tokens, their relative start state, impact-related annotations, and the average transition variable:
\begin{equation}
\mathbf{z}^{\mathrm{pair}}_{ij}
=
\Psi_{\mathrm{pair}}
\!\left(
\mathbf{z}_i,
\mathbf{z}_j,
\mathbf{p}_{j,0}-\mathbf{p}_{i,0},
\mathbf{q}^{\mathrm{side}}_{ij},
\mathbf{q}^{\mathrm{init}}_{i},
\mathbf{q}^{\mathrm{init}}_{j},
\frac{\hat{\tau}_i+\hat{\tau}_j}{2}
\right).
\end{equation}

This pair token is broadcast across time and processed by a temporal pair graph:
\begin{equation}
\mathbf{g}^{(\ell+1)}_{ij,k}
=
\mathrm{GT}^{(\ell)}_{\mathrm{pair}}
\!\left(
\mathbf{g}^{(\ell)}_{ij,k},
\mathcal{N}_{\mathrm{pair}}(k),
\mathbf{e}^{\mathrm{pair}}_{k\rightarrow k'}
\right).
\end{equation}

The updated pair states are decoded into relative-motion corrections,
\begin{equation}
\Delta \mathbf{r}_{ij,k}
=
\Psi_{\mathrm{pair\_out}}(\mathbf{g}^{(L)}_{ij,k}),
\end{equation}
yielding the refined pair trajectory
\begin{equation}
\hat{\mathbf{r}}_{ij,k}
=
\mathbf{r}^{\mathrm{base}}_{ij,k}
+
\sum_{\kappa=1}^{k}
\Delta \mathbf{r}_{ij,\kappa}.
\end{equation}

The pairwise refinement is then injected back into the scene through a post-transition fusion module. Let \(\Delta \mathbf{o}_{i,k}\) and \(\Delta \mathbf{o}_{j,k}\) denote the fusion offsets predicted for the target pair. Using a sigmoid gate centered on the average transition time,
\begin{equation}
g_{ij,k}
=
\sigma\!\left(
\lambda_{\tau}
\left(
t_k-\frac{\hat{\tau}_i+\hat{\tau}_j}{2}
\right)
\right),
\end{equation}
the fused trajectories are written as
\begin{equation}
\begin{split}
\mathbf{p}^{\mathrm{fuse}}_{i,k}
=
\mathbf{p}^{\mathrm{base}}_{i,k}
+
g_{ij,k}\Delta \mathbf{o}_{i,k},
\\
\mathbf{p}^{\mathrm{fuse}}_{j,k}
=
\mathbf{p}^{\mathrm{base}}_{j,k}
+
g_{ij,k}\Delta \mathbf{o}_{j,k}.
\end{split}
\end{equation}

Finally, the annotated accident location provides a weak scene-level geometric anchor. Let
\begin{equation}
\bar{\mathbf{p}}^{\mathrm{end}}
=
\frac{1}{|\mathcal{I}_{\mathrm{valid}}|}
\sum_{i\in\mathcal{I}_{\mathrm{valid}}}
\mathbf{p}^{\mathrm{fuse}}_{i,K}
\end{equation}
denote the mean terminal position of all valid vehicles, and let
\begin{equation}
\Delta \mathbf{a}
=
\mathbf{a}^{\ast}
-
\bar{\mathbf{p}}^{\mathrm{end}}
\end{equation}
denote the offset from the reconstructed terminal configuration to the annotated accident location \(\mathbf{a}^{\ast}\). This offset is softly injected after the transition boundary:
\begin{equation}
\mathbf{p}^{\mathrm{geom}}_{i,k}
=
\mathbf{p}^{\mathrm{fuse}}_{i,k}
+
\sigma\!\bigl(
\lambda_{a}(t_k-\hat{\tau}_i)
\bigr)\Delta \mathbf{a}.
\end{equation}

As a result, the geometric decoder produces scene-grounded trajectories that remain lane consistent at the global level while preserving collision relevant interaction at the local level.

\subsection{Temporal allocation under physical consistency}

After geometric reconstruction, the timing head allocates motion along the reconstructed path so that the final trajectories remain dynamically plausible under report-derived behavioral and speed-related cues. Rather than regenerating geometry, this module re-parameterizes motion along the scene-grounded paths produced by the geometric decoder. This design disentangles where each vehicle moves from how quickly it traverses the reconstructed path, which is particularly important when geometric evidence is stronger than temporal supervision.

For vehicle \(i\), let \(\mathbf{p}^{\mathrm{geom}}_{i,k}\) denote the geometrically reconstructed trajectory at time step \(k\). The corresponding geometric step length is defined as
\begin{equation}
d^{\mathrm{geom}}_{i,k}
=
\left\|
\mathbf{p}^{\mathrm{geom}}_{i,k}
-
\mathbf{p}^{\mathrm{geom}}_{i,k-1}
\right\|_2.
\end{equation}
Conditioned on vehicle-level semantic descriptors, speed-related cues, and the transition variable \(\hat{\tau}_i\), the timing head predicts a residual temporal adjustment factor \(\delta_{i,k}\), which modulates the geometric step length as
\begin{equation}
\tilde{d}_{i,k}
=
d^{\mathrm{geom}}_{i,k}\exp(\delta_{i,k}).
\end{equation}

To preserve motion plausibility, the adjusted step length is further regularized by soft speed and smoothness constraints:
\begin{equation}
d^{\mathrm{final}}_{i,k}
=
\Phi_{\mathrm{phys}}
\!\left(
\tilde{d}_{i,k};
\mathbf{b}_i,
\mathbf{u}^{\mathrm{spd}}_i,
\hat{\tau}_i
\right),
\end{equation}
where \(\Phi_{\mathrm{phys}}(\cdot)\) denotes a physics-aware temporal adjustment operator that suppresses implausible speed excursions and abrupt local fluctuations while preserving consistency with the available motion cues.

The resulting cumulative arc length along the reconstructed path is then written as
\begin{equation}
\ell_{i,k}
=
\sum_{\kappa=1}^{k}
d^{\mathrm{final}}_{i,\kappa}.
\end{equation}
The final pre-impact trajectory is obtained by arc-length resampling of the geometric path:
\begin{equation}
\hat{\mathbf{p}}_{i,k}
=
\mathrm{ArcResample}
\!\left(
\mathbf{p}^{\mathrm{geom}}_{i,:},
\ell_{i,k}
\right).
\end{equation}
The corresponding speed is recovered from adjacent displacements,
\begin{equation}
\hat{v}_{i,k}
=
\frac{
\left\|
\hat{\mathbf{p}}_{i,k}
-
\hat{\mathbf{p}}_{i,k-1}
\right\|_2
}{\Delta t},
\qquad
\Delta t = 0.1~\mathrm{s}.
\end{equation}

As a result, the final reconstruction preserves both the geometric structure inferred from scene-grounded accident reasoning and the temporal evolution required for behavior-consistent and physically plausible pre-impact motion.

\subsection{Loss function}

We optimize the model using a structured objective that emphasizes trajectory fidelity, collision relevant interaction, temporal consistency, and motion plausibility. The primary supervision is imposed on the final reconstructed motion, the localized collision geometry, and the speed profile, while additional auxiliary terms are used only to stabilize training and improve structural consistency.

The primary geometric supervision is given by the trajectory reconstruction loss, which constrains the final reconstructed trajectory \(\hat{\mathbf{p}}_{i,k}\) to match the dense reference trajectory \(\mathbf{p}^{\ast}_{i,k}\) on all valid vehicle--time pairs:
\begin{equation}
\mathcal{L}_{\mathrm{traj}}
=
\frac{1}{|\Omega|}
\sum_{(i,k)\in\Omega}
\mathrm{SmoothL1}
\!\left(
\hat{\mathbf{p}}_{i,k},
\mathbf{p}^{\ast}_{i,k}
\right),
\end{equation}
where \(\Omega\) denotes the set of valid spatiotemporal states.

To preserve the geometry of collision relevant interaction, we further impose a pairwise interaction loss on the selected target collision pair \((i,j)\):
\begin{equation}
\mathcal{L}_{\mathrm{pair}}
=
\frac{1}{K}
\sum_{k=1}^{K}
\left\|
\hat{\mathbf{r}}_{ij,k}
-
\mathbf{r}^{\ast}_{ij,k}
\right\|_{1},
\end{equation}
where \(\hat{\mathbf{r}}_{ij,k}\) and \(\mathbf{r}^{\ast}_{ij,k}\) denote the predicted and reference relative motion, respectively. This term directly constrains the localized interaction geometry that governs approach, evasive deviation, and collision formation.

Beyond relative motion, accident reconstruction also requires consistency with the annotated collision configuration. We therefore introduce a collision consistency loss,
\begin{equation}
\mathcal{L}_{\mathrm{coll}}
=
\phi_{\mathrm{coll}}
\!\left(
\hat{\mathbf{p}}_{i,:},
\hat{\mathbf{p}}_{j,:},
\mathbf{a}^{\ast}
\right),
\end{equation}
where \(\phi_{\mathrm{coll}}(\cdot)\) penalizes incompatibility between the reconstructed target-pair motion and the annotated accident location \(\mathbf{a}^{\ast}\) in the vicinity of collision.

Temporal consistency is supervised through a speed fitting loss on reliable speed annotations:
\begin{equation}
\mathcal{L}_{\mathrm{spd}}
=
\frac{1}{|\Omega_v|}
\sum_{(i,k)\in\Omega_v}
\left|
\hat{v}_{i,k}
-
v_{i,k}
\right|,
\end{equation}
where \(\Omega_v\) denotes the set of valid speed-supervision steps.

% To preserve motion plausibility when direct supervision is incomplete or noisy, we additionally apply a physics-aware regularization term:
% \begin{equation}
% \mathcal{L}_{\mathrm{phys}}
% =
% \mathcal{L}_{\mathrm{beh}}
% +
% \mathcal{L}_{\mathrm{limit}}
% +
% \mathcal{L}_{\mathrm{smooth}},
% \end{equation}
% where \(\mathcal{L}_{\mathrm{beh}}\) penalizes acceleration patterns that are inconsistent with the reported behavioral cues, \(\mathcal{L}_{\mathrm{limit}}\) suppresses implausible speed violations relative to the available speed-related constraints, and \(\mathcal{L}_{\mathrm{smooth}}\) discourages abrupt local motion changes, including unreasonable acceleration/deceleration and curvature changes.

To ensure that the temporally allocated trajectories remain dynamically plausible, we impose a compact physical-consistency regularization on the reconstructed motion,
\begin{equation}
\mathcal{L}_{\mathrm{phys}}
=
\mathcal{L}_{\mathrm{beh}}
+
\mathcal{L}_{\mathrm{limit}}
+
\mathcal{L}_{\mathrm{smooth}}.
\end{equation}
Rather than enforcing hard kinematic rules, these terms act as soft priors on behavior consistency, speed reasonableness, and local motion continuity.

For behavior consistency, we constrain the post-transition acceleration trend to remain compatible with the annotated avoidance maneuver. Let \(a_t\) denote the discrete acceleration and let \(\Omega_{\mathrm{post}}\) be the valid post-transition interval. We compute the mean post-transition acceleration
\begin{equation}
\bar a_{\mathrm{post}}
=
\frac{1}{|\Omega_{\mathrm{post}}|}
\sum_{t \in \Omega_{\mathrm{post}}} a_t,
\end{equation}
and define \(\mathcal{L}_{\mathrm{beh}}\) according to the behavior label: acceleration-related behaviors are encouraged to have positive \(\bar a_{\mathrm{post}}\), braking-related behaviors are encouraged to have negative \(\bar a_{\mathrm{post}}\), whereas neutral behaviors are regularized to avoid unnecessary acceleration fluctuations. In this way, the reconstructed post-event motion remains semantically aligned with the reported avoidance response.

For speed reasonableness, we introduce a soft speed-limit constraint based on the reported speed prior \(s_{\mathrm{lim}}\). Instead of enforcing a hard upper bound, we penalize only excessive overspeeding beyond a tolerance margin \(\delta\),
\begin{equation}
\mathcal{L}_{\mathrm{limit}}
=
\frac{1}{|\Omega_v|}
\sum_{t \in \Omega_v}
\operatorname{SmoothL1}
\!\left(
\max\!\bigl(0,\, v_t - (s_{\mathrm{lim}}+\delta)\bigr),\, 0
\right),
\end{equation}
where \(v_t\) is the reconstructed speed and \(\Omega_v\) denotes the valid time steps. This term suppresses unrealistic high-speed motion while retaining flexibility for accident scenarios that may involve moderate speed deviation.

For local motion continuity, we regularize abrupt temporal changes in motion direction. Let \(\Delta \mathbf{p}_t = \mathbf{p}_t - \mathbf{p}_{t-1}\) denote the displacement between adjacent steps, and let
\begin{equation}
c_t
=
\frac{\Delta \mathbf{p}_t^{\top}\Delta \mathbf{p}_{t+1}}
{\|\Delta \mathbf{p}_t\|_2 \, \|\Delta \mathbf{p}_{t+1}\|_2}
\end{equation}
be the cosine similarity between consecutive motion directions. The smoothness term is then written as
\begin{equation}
\mathcal{L}_{\mathrm{smooth}}
=
\frac{1}{|\Omega_d|}
\sum_{t \in \Omega_d}
\max\!\left(0,\, \cos\theta_0 - c_t\right),
\end{equation}
where \(\Omega_d\) contains valid non-stationary time steps and \(\theta_0\) is the maximum turning angle allowed without penalty. This term discourages implausible high-frequency oscillation and overly sharp local turns, thereby improving temporal coherence.

Together, these three components provide weak but effective physical priors that guide the reconstructed trajectories toward behavior-consistent, speed-reasonable, and temporally smooth pre-impact motion.

% In addition to the primary objectives, several auxiliary supervision terms are used to stabilize optimization and improve structural consistency. These include lane-candidate supervision, heading consistency, transition-time supervision, and branch-level regularization:
% \begin{equation}
% \mathcal{L}_{\mathrm{aux}}
% =
% \mathcal{L}_{\mathrm{cand}}
% +
% \mathcal{L}_{\mathrm{head}}
% +
% \mathcal{L}_{\tau}
% +
% \mathcal{L}_{\mathrm{reg}},
% \end{equation}
% where \(\mathcal{L}_{\mathrm{cand}}\) supervises lane-candidate selection, \(\mathcal{L}_{\mathrm{head}}\) enforces local heading consistency, \(\mathcal{L}_{\tau}\) constrains the predicted transition variable, and \(\mathcal{L}_{\mathrm{reg}}\) collects additional branch-level and auxiliary regularization terms.

The overall training objective is therefore written as
\begin{equation}
% \mathcal{L}
% =
% \mathcal{L}_{\mathrm{traj}}
% +
% \lambda_{\mathrm{pair}}\mathcal{L}_{\mathrm{pair}}
% +
% \lambda_{\mathrm{coll}}\mathcal{L}_{\mathrm{coll}}
% +
% \lambda_{\mathrm{spd}}\mathcal{L}_{\mathrm{spd}}
% +
% \lambda_{\mathrm{phys}}\mathcal{L}_{\mathrm{phys}}
% +
% \mathcal{L}_{\mathrm{aux}}.
\mathcal{L}
=
\mathcal{L}_{\mathrm{traj}}
+
\mathcal{L}_{\mathrm{pair}}
+
\mathcal{L}_{\mathrm{coll}}
+
\mathcal{L}_{\mathrm{spd}}
+
\mathcal{L}_{\mathrm{phys}}
\end{equation}

Taken together, these objectives supervise accident reconstruction from complementary but aligned perspectives. The primary terms enforce trajectory fidelity, collision relevant interaction, temporal consistency, and motion plausibility, whereas the auxiliary terms mainly improve optimization stability and structural coherence.

\begin{figure*}[t]
\centering 
\includegraphics[width=0.90\textwidth]{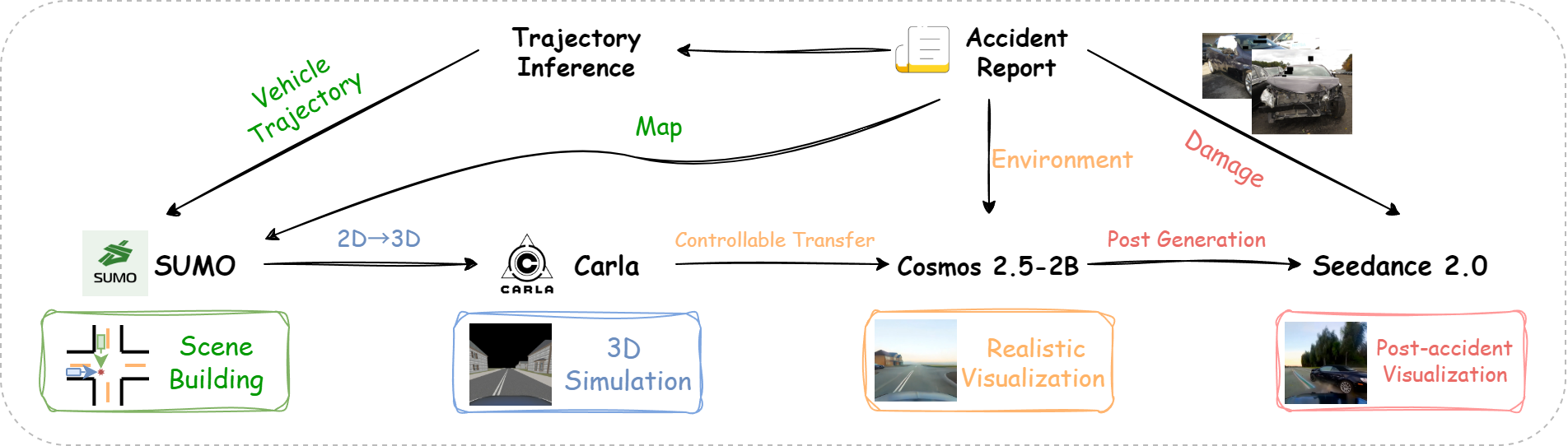}
\caption{Controllable visual reconstruction of traffic accident scenes.
The framework first uses traffic simulation to preserve the physical consistency of scene geometry, vehicle motion and participant interactions, and then applies AI-based video generation to translate simulator outputs into realistic accident visualizations with matched environmental conditions and post-collision appearance.}
\label{visual}
\end{figure*}

\subsection{Controllable visual rendering}

To facilitate qualitative analysis of the reconstructed motion and scene observation, we further couple the inferred trajectories with a downstream controllable rendering pipeline that converts scene-grounded accident reconstruction into visually interpretable accident replay. This module renders the reconstructed trajectories within a scene-consistent visual environment and provides a controllable interface for examining accident dynamics from multiple viewpoints and under different contextual conditions, as illustrated in Figure~\ref{visual}.

The rendering pipeline consists of three stages: scene instantiation, visual style transfer, and post-impact extension. First, the reconstructed scene geometry and vehicle trajectories are instantiated in a simulation environment. We use SUMO~\cite{krajzewicz2012recent} to obtain or construct the street-level map associated with the accident location, and then import the reconstructed trajectories together with the corresponding road layout into CARLA. This stage converts the reconstructed 2D scene representation into a simulation-ready 3D environment, enabling scene-consistent visualization of the inferred pre-impact dynamics.

Second, the simulator-style rendering is transferred into a more realistic visual domain. Specifically, we employ Cosmos-Transfer2.5-2B~\cite{ali2025world} to convert the rendered outputs into visually richer accident scenes while allowing environmental attributes such as weather, illumination, and scene appearance to be adjusted in a controllable manner. In this way, the rendered videos preserve consistency with the reconstructed scene layout and vehicle motion while providing more interpretable visual presentation. f

Because the core reconstruction model focuses on the pre-impact phase, post-impact visualization is treated as a downstream generative extension rather than as a supervised reconstruction target. To extend the rendered results beyond the moment of collision, we further condition on damage-related descriptions and post-impact behavioral cues, and use Seedance~2.0 to generate post-collision visualizations. This extension is intended to provide qualitative visualization of accident aftermath, including vehicle damage manifestation and post-impact scene evolution, rather than quantitatively supervised accident reconstruction.

\begin{figure*}[t]
\centering 
\includegraphics[width=0.95\textwidth]{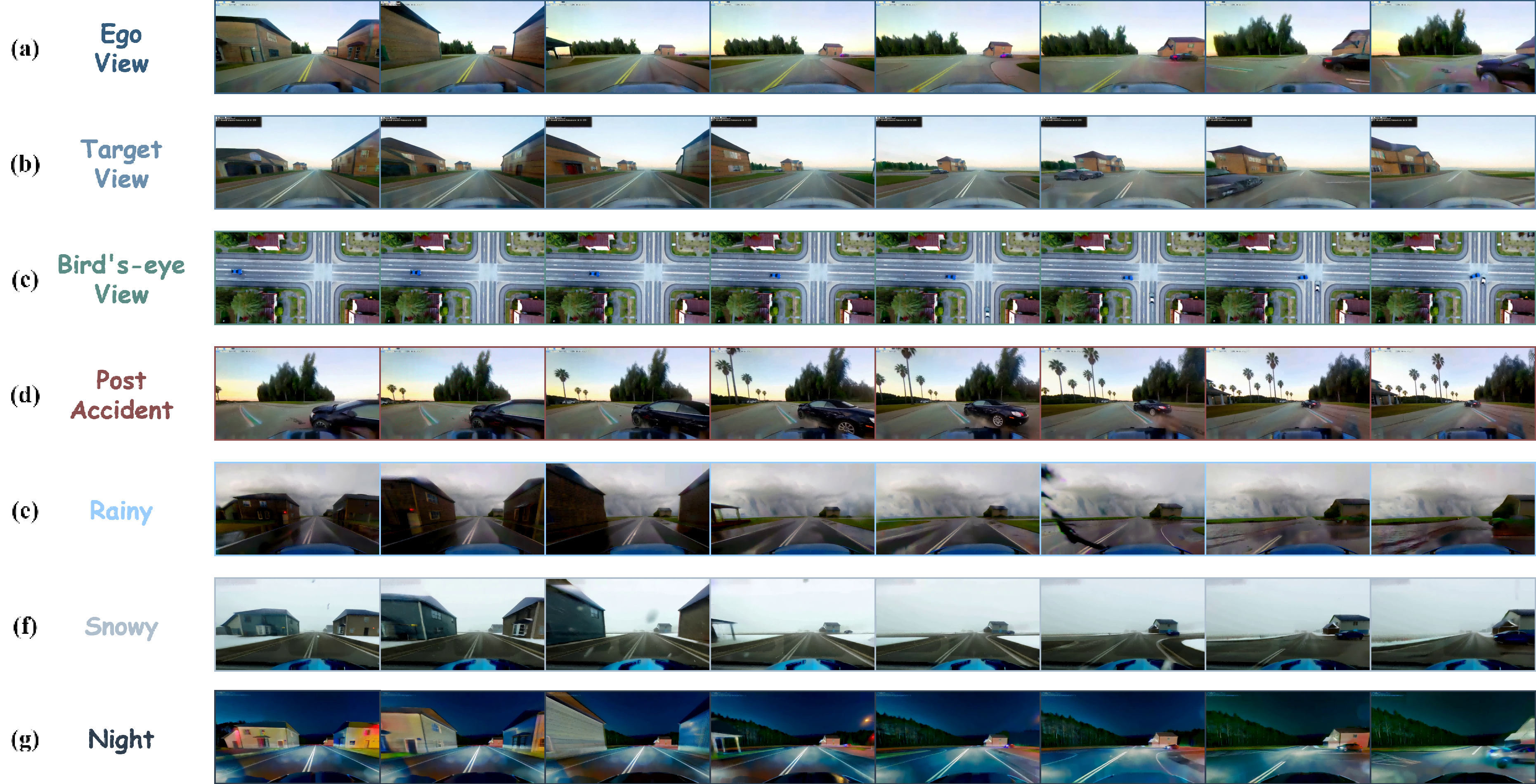}
\caption{Controllable visualization results of a reconstructed traffic accident case, including multi-view pre-impact replay, post-impact scene extension, and rendering under different environmental conditions.}
\label{example}
\end{figure*}

Figure~\ref{example} illustrates the realistic visualization results of a single traffic accident case, including multi-view comparisons, post-impact reconstruction, and visual renderings under different environmental conditions. Specifically, Figure~\ref{example}(a) presents the accident scene from the ego-vehicle dashcam perspective, while Figure~\ref{example}(b) shows the viewpoint of the opposing vehicle involved in the Figure~\ref{example}(c) provides a BEV visualization of the accident scene, offering an objective and comprehensive overview of the spatial relationships and vehicle interactions.

Figure~\ref{example}(d) depicts the post-impact scene generated according to the described vehicle damage conditions, reconstructing the accident aftermath after the collision. As shown in the figure, the hood of the ego vehicle is lifted due to frontal impact, while the side door of the target vehicle is deformed, reflecting the collision geometry and impact severity. Figure~\ref{example}(e--g) present video style transfer results under different environmental conditions, demonstrating controllable rendering of diverse accident contexts.

It can be observed that across all generated videos, the road structure and vehicle behaviors remain consistent, strictly following the reconstructed map and inferred trajectories. The multi-view renderings mutually validate the spatial consistency of the reconstruction. Meanwhile, controllable environmental rendering aims to faithfully reproduce the contextual conditions at the time of the accident. As a result, the final outputs achieve strong consistency with real-world traffic accidents in terms of both physical structure and visual realism.

\section{Experiments}

\subsection{Dataset}

\begin{figure*}[t]
\centering 
\includegraphics[width=0.88\textwidth]{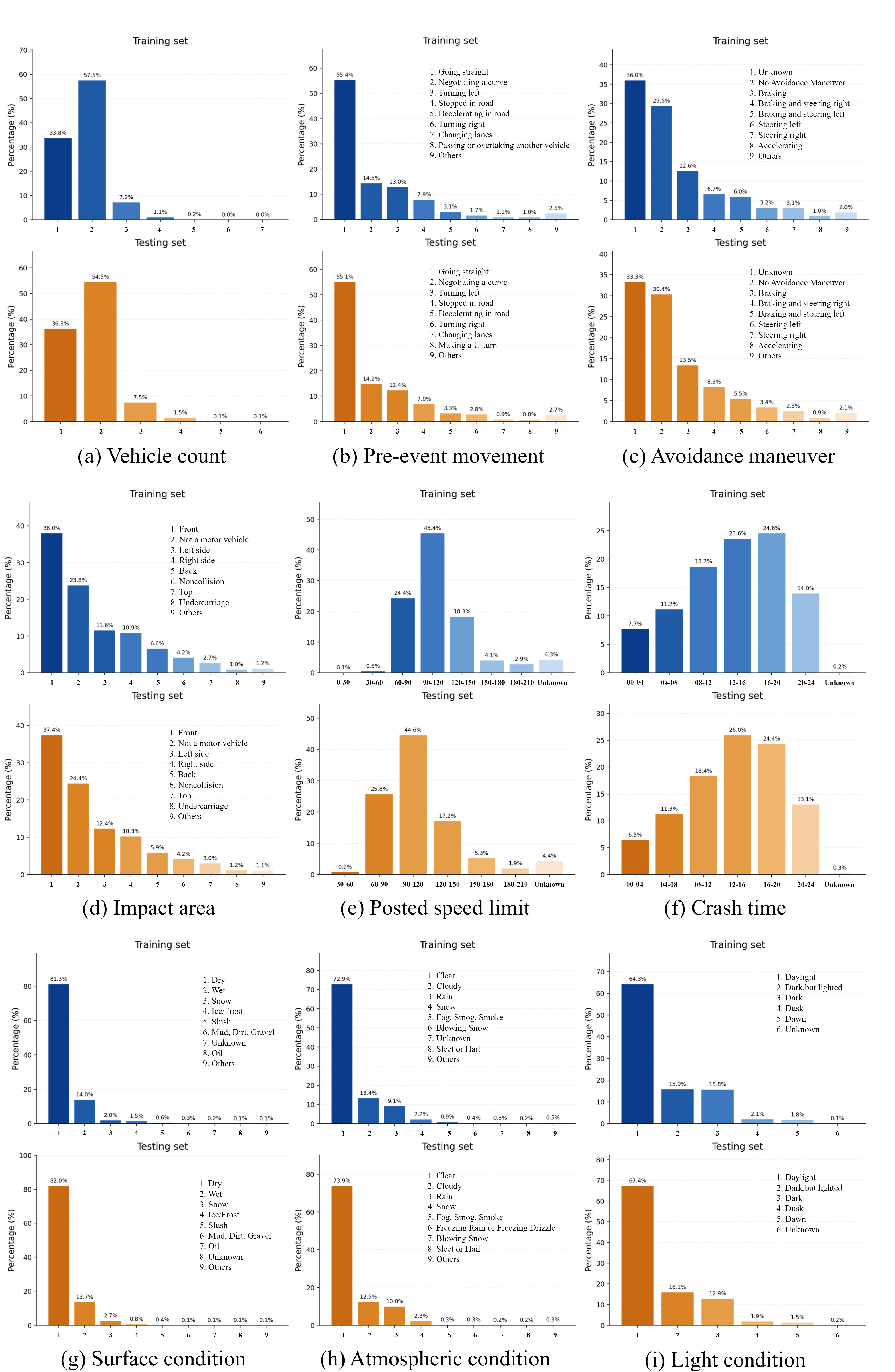}
\caption{Data distribution of key factors in the CISS-REC dataset.}
\label{dataset}
\end{figure*}

\begin{table*}[t]
\centering
\caption{Noise analysis of key input information in CISS-REC. The reported values correspond to the percentage of vehicles in the subset for which the corresponding fields are missing, labeled as unknown or contain content but do not conform to the expected format.}
\begin{tabular}{c c c c c c c c c}
\hline
Subset& Source &Trajectory & Impact Area & Speed Limit & Pre-movement & Avoidance& EDR Data & Initial Lane\\
\hline
\multirow{3}{*}{Training}& Missing& 5.98 & 0 & 2.97 & 2.97 & 2.97 & 59.03 &2.97 \\
      &Unknown& - & 0.95 & 0.75 & 0.07 & 34.95 & 0 &0.15 \\
      &Error& - & 0      & 0.62 & 0 & 0 & 0 &0.05 \\
\hline 
\multirow{3}{*}{Testing}& Missing& 5.89 & 0 & 3.32 & 3.32 & 3.32 & 57.09 &3.32 \\
       &Unknown& - & 0.80 & 0.97 & 0 & 32.21 & 0 &0.06 \\
      &Error& - & 0 & 0.11 & 0 & 0 & 0 &0.06 \\
\hline 
\end{tabular}
\label{noise}
\end{table*}

To promote the sustainability and reproducibility of research in this domain, we focus on processing accident reports derived from publicly accessible data sources and release a corresponding accident reconstruction dataset CISS-REC. Specifically, we curate accident reports from the NHTSA CISS and perform data preprocessing as described in Section~\ref{preprocessingtex}. During the data preparation process, we exclude cases in which the accident sketches were generated through simple manual drawing using tools such as Google Maps. Instead, we retain only those cases in which the accident scenes were measured on site using total station surveying equipment and documented with sketches that strictly follow standardized naming conventions.

This selection strategy aims to eliminate inconsistencies introduced by variations in investigation workflows and heterogeneous labeling or naming systems across accident reports. By restricting the dataset to cases with precise surveying measurements and standardized documentation, we ensure a higher level of geometric accuracy and structural consistency in the reconstructed accident scenes.

After filtering, the resulting dataset contains a total of 6{,}217 accident cases. Among them, 1{,}000 cases are randomly selected as the test set, while the remaining cases are used for training. This dataset provides a structured foundation for studying parameterized accident reconstruction and for evaluating the performance of machine-learning-based reconstruction models under realistic conditions.

As shown in Fig.~\ref{dataset}, the training and testing sets follow closely matched distributions across vehicle composition, participant behavior, impact configuration, environmental condition, and temporal context, suggesting that the random split preserves the overall statistical structure of CISS-REC. Across both subsets, two-vehicle crashes form the dominant scenario type, while crashes involving more participants are progressively less frequent. Pre-event movement is primarily characterized by going straight, negotiating a curve, and turning left, whereas more complex behaviors, such as lane changing, overtaking, and U-turn-related maneuvers, occur only sparsely.

A similarly skewed pattern is observed for avoidance maneuvers and impact areas. Unknown or absent avoidance annotations account for a substantial proportion, followed by no maneuver and braking-related responses, while coupled steering-and-braking actions remain relatively uncommon. Front-impact collisions constitute the largest category, whereas rear and lateral impact configurations are more weakly represented. These observations indicate that the dataset reflects the intrinsic long-tailed nature of accident reports, where common traffic interactions dominate but rare yet safety-critical patterns are still retained.

Environmental distributions further show that most cases occur on dry roads, under clear weather, and during daylight, with smaller yet non-negligible proportions under wet, rainy, dark, or dusk conditions. Posted speed limits are mainly concentrated in the medium-speed intervals, and crash occurrence is more frequent during daytime than overnight periods. In addition, about 75\% of accidents occur on straight roads, while the proportions for left-turn and right-turn roads are about 15\% and 10\%, respectively. Taken together, these results demonstrate that CISS-REC provides a realistic and statistically coherent benchmark for accident reconstruction, while preserving the class imbalance and contextual heterogeneity inherent to real-world in-depth crash data.

Table~\ref{noise} presents the noise distribution of key attributes in the CISS-REC dataset. These noise sources mainly originate from missing entries in the original records as well as data entry errors during the documentation process. For example, missing trajectory information is primarily caused by inaccuracies in accident sketch drawing and annotation. In contrast, noise in objective attributes such as speed limits mainly arises from incomplete data collected during the investigation process.

Notably, attributes related to driver intent and those requiring specialized equipment exhibit a relatively high proportion of missing or unknown values. This is largely due to privacy protection considerations, the varying depth of accident investigation procedures, and limitations in vehicle instrumentation. As a result, certain fields are unavailable for a partial subset of cases. Therefore, during the accident reconstruction process, it is necessary to selectively utilize valid and reliable information for supervised learning, ensuring that model training is guided by consistent and informative inputs.

\subsection{Evaluation metrics}
\label{Evaluation metrics}

To quantitatively evaluate the discrepancy between the reconstructed traffic accident scenes and the scenarios described in the original accident reports, we define the following evaluation metrics.

\paragraph{Average Keypoint Distance (AKD).}
This metric measures the spatial deviation of each traffic participant at critical trajectory points mapped in the reconstructed scene. Specifically, for each participant involved in the accident, we compute the Euclidean distance between the reconstructed position and the ground-truth position at keypoints obtained from the surveyed accident report. The final score is obtained by averaging the distances across all participants and all keypoints,
\begin{equation}
AKD = \frac{1}{N K} 
\sum_{i=1}^{N} \sum_{k=1}^{K}
\left\| \mathbf{p}^{rec}_{i,k} - \mathbf{p}^{gt}_{i,k} \right\|_2,
\end{equation}
where $N$ is the number of traffic participants, $K$ is the number of keypoints, and $\mathbf{p}_{i,k}$ denotes the position of participant $i$ at keypoint $k$.

\paragraph{Average Velocity Deviation (AVD).}
This metric quantifies the difference between the reconstructed motion states and the original EDR measurements. For each participant and each timestamp specified in the EDR data, the absolute error between the reconstructed velocity and the reference velocity is calculated. The errors are then averaged over all timestamps and all participants. This metric primarily reflects the capability of the reconstruction model to capture the motion intensity, driving speed, and dynamic evolution of the accident process,
\begin{equation}
AVD = \frac{1}{N T}
\sum_{i=1}^{N} \sum_{t=1}^{T}
\left| V^{rec}_{i,t} - V^{gt}_{i,t} \right|,
\end{equation}
where $T$ is the number of evaluated timestamps.

\paragraph{Average Accident-Point Distance ($AAPD_{tan \& norm}$).}
The accident-point distance specifically evaluates the accuracy of the reconstructed collision location. Considering that a single Euclidean distance cannot fully capture the semantic implications of collision errors, we decompose the positional error into two orthogonal components: the tangential error along the road direction (or the principal motion direction of the vehicle) and the normal error perpendicular to that direction. The tangential distance reflects the longitudinal displacement of the collision point, indicating whether the collision occurred ``too early'' or ``too late'' along the trajectory. The normal distance measures the lateral deviation of the collision position, indicating whether the collision occurred in the correct lane and whether the relative lateral alignment between vehicles is reasonable. This decomposition is more suitable for traffic accident scenarios because it distinguishes between temporal misalignment and lateral localization errors,
\begin{equation}
\begin{aligned}
AAPD_{tan} &=
\left| (\mathbf{c}^{rec}-\mathbf{c}^{gt}) \cdot \mathbf{e}_{tan} \right|, \\
AAPD_{norm} &=
\left| (\mathbf{c}^{rec}-\mathbf{c}^{gt}) \cdot \mathbf{e}_{norm} \right|,
\end{aligned}
\end{equation}
where $\mathbf{c}$ denotes the collision point and $\mathbf{e}_{tan}, \mathbf{e}_{norm}$ represent the unit vectors along the tangential and normal directions respectively.

\paragraph{Collision Rate (CR).}
The collision rate measures whether the reconstructed multi-vehicle scene successfully reproduces the collision event described in the original accident report. For accident samples where a collision is known to occur, if the reconstructed trajectories of the involved participants satisfy a predefined collision detection condition, the reconstruction is considered successful; otherwise, it is regarded as a failure. Assuming typical passenger vehicle dimensions (approximately $15$~ft in length and $6$~ft in width), the conservative circular approximation yields a geometric contact condition when the center distance between two vehicles is less than or equal to $16.2$~ft. In our implementation, we adopt the larger value between $15$~ft and the original collision distance recorded in the report as the collision detection threshold.

\paragraph{Collision Surface Accuracy (CSA).}
Collision surface accuracy evaluates whether the contact surfaces involved in the reconstructed collision match those described in the accident report. This metric assesses the correctness of the collision topology and the geometric semantics of the contact region, providing a finer-grained evaluation of collision reconstruction quality.

\paragraph{Behavior Accuracy (BA).}
Behavior accuracy evaluates whether the model correctly reconstructs the key actions or driving behaviors of each participant before and during the accident. During evaluation, LLMs are used to analyze the reconstructed driving behaviors and evasive maneuvers of the participants. The inferred behaviors are then compared with those described in the original data to calculate the overall accuracy.

\paragraph{Role Accuracy (RA).}
Role accuracy measures whether the model correctly recovers the functional or causal roles of traffic participants in the accident. In traffic accident analysis, different participants may assume distinct roles, and the identification of the \emph{initiator} is particularly important for determining accident responsibility. In our evaluation, LLMs are employed to assist in identifying the roles of reconstructed participants. The predicted roles are then compared with the roles specified in the original accident report to compute classification accuracy.

% \paragraph{Semantic Similarity (SS).}
% Semantic similarity provides a higher-level assessment of the consistency between the reconstructed scene and the original accident report. Specifically, the reconstructed accident sketch and the textual accident report are encoded into a shared semantic embedding space, and their similarity is computed within that space. Unlike purely geometric metrics, this measure focuses on whether the reconstructed scene preserves the core semantic information of the original accident report at the event level.

\begin{table*}[t]
\centering
\caption{Quantitative comparison of different models for text-based conditional inverse inference on partially observed multi-agent trajectories. Distance-based metrics are reported in $m$, velocity-based metrics in $m/s$, and accuracy-based metrics in \%. }
\resizebox{0.98\linewidth}{!}{
\begin{tabular}{c c c cc c c c c c c c}
\hline
\multirow{2}{*}{Model} & \multirow{2}{*}{AKD↓} & \multirow{2}{*}{AVD} & \multicolumn{2}{c}{AAPD} & \multirow{2}{*}{CR} & \multirow{2}{*}{CSA} & \multirow{2}{*}{BA} & \multirow{2}{*}{RA} & \multirow{2}{*}{Acc} & \multirow{2}{*}{Curvature} \\
\cline{4-5}
& & & tan & norm & & & & \\
\hline

STGCN~\cite{han2020stgcn}& 12.15 & 4.82 & 9.09 & 3.71 & 19.70 & 28.79 & 4.95 & 74.73 & 25.35 & 1.03\\
Spline~\cite{schumaker2007spline}& 12.01 & 7.42 & 16.57 & 6.21 & 22.73 & 53.73 & 3.30 & 61.54 &14.44  &0.18\\
 LSTM~\cite{van2020review} & 11.88 & 10.85 &9.28  &4.68  &16.67  &41.79  & 40.11 & 76.81 & 4.10 &0.43\\
  Wayformer~\cite{nayakanti2022wayformer}& 11.64 & 5.72 & 8.83 & 3.52 & 22.73 & 32.58 &5.49  & 75.27 &39.32  &1.08\\
 Spline with temporal allocation & 11.35 & 6.62 & 8.06 & 3.45 & 21.21 & 44.78 &2.20  & 59.34 & 11.71 &0.20\\
HiVT~\cite{zhou2022hivt}& 11.34 & 6.12 & 8.54 & 3.56 & 19.70 & 41.04 & 1.65 & 72.53 & 51.65 &1.05\\
Kinematic fitting & 7.80 & 8.90 & 9.49 & 5.41 & 22.73 & 20.45 &42.86  & 79.01 & 3.67 & 0.02\\
 VectorNet~\cite{gao2020vectornet} & 7.63 & 6.36 & 6.37 & 3.82 & 40.91 & 26.87 &40.11  & 87.91 & 33.40 & 0.89\\
PC-Crash~\cite{steffan1996collision} & 6.29 & 9.53 & 5.31 & 3.29 & 98.48 & 8.96 &37.91  & 84.07 & 21.81 &0.14\\
Momentum-Energy~\cite{zhang2006virtual} & 6.09 & 13.81 & 5.63 & 3.43 & 100.00& 17.91 &17.58  & 73.08 & 28.46 &0.05\\

 \hline 
 Our & 4.98 & 6.84 & 3.47 & 1.41 &85.71 & 56.35 & 59.69 & 87.88 & 2.23 & 0.04\\
\hline 
\end{tabular}
}
\label{results}
\end{table*}

\begin{figure*}[t]
\centering 
\includegraphics[width=0.95\textwidth]{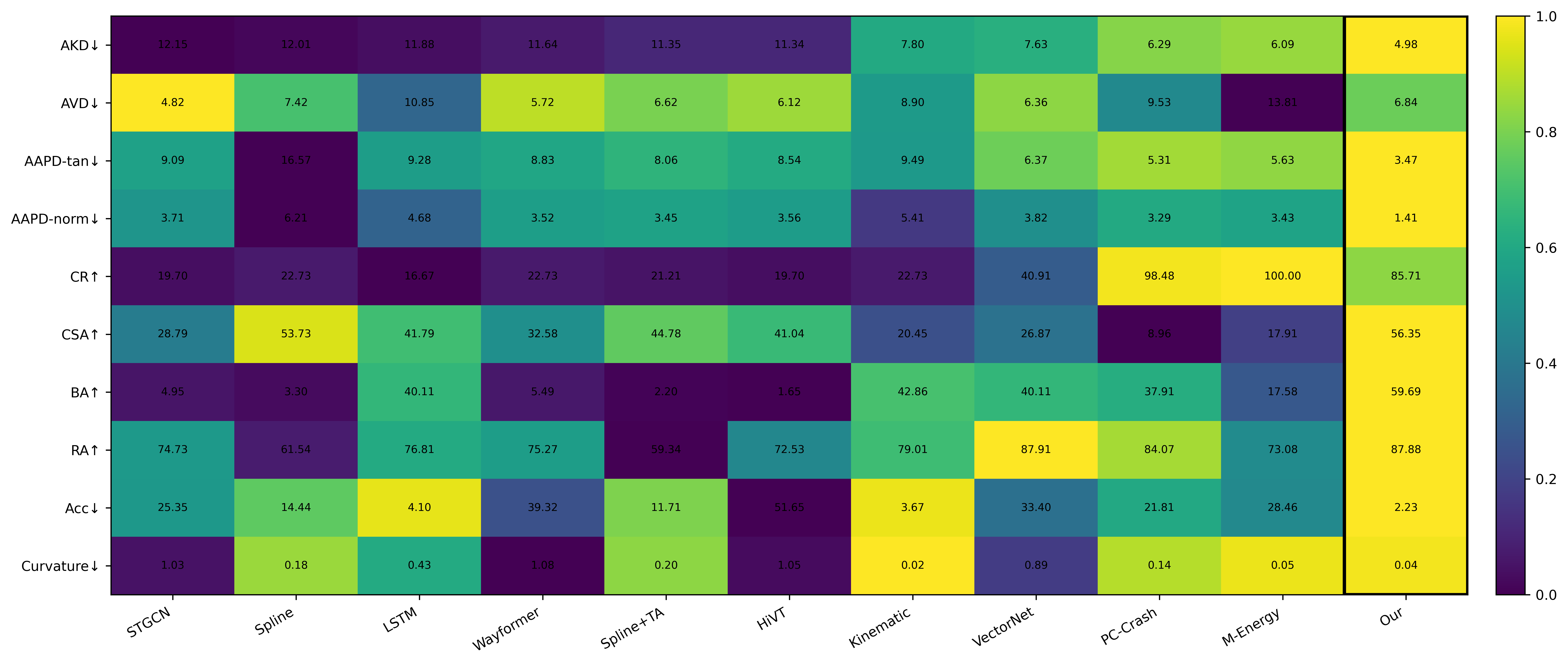}
\caption{Metric landscape of accident reconstruction performance across baselines. Higher scores represent better performance across trajectory, collision, behavior, and geometry metrics.}
\label{result_hot}
\end{figure*}

\subsection{Experimental setups}

All experiments are conducted on the preprocessed multimodal accident dataset described above. Each sample contains road geometry, scene- and vehicle-level textual embeddings, structured behavioral attributes, accident priors, dense trajectory supervision, and reliability masks. Unless otherwise specified, each scene contains up to five participant slots, the prediction horizon is fixed to $K=51$ time steps covering the 5~s interval before impact, and all trajectories are represented in the standardized north-up coordinate frame.

The model is implemented in PyTorch. The latent feature dimension follows the default architecture configuration used in the released implementation, and the encoder, base decoder, pairwise interaction branch, and timing head are trained jointly under automatic mixed precision when CUDA is available. Optimization is performed using AdamW with grouped learning rates, so that the encoder is updated more conservatively than the newly introduced conditional decoding modules. Gradient clipping is applied throughout training to maintain numerical stability, and batches that produce non-finite losses are skipped during optimization.

Rather than optimizing the full architecture end-to-end from the outset, we adopt a three-stage training strategy. In the first stage, only the base decoder is trained while the encoder remains frozen, allowing the model to first learn stable lane-aware trajectory reconstruction and transition-time estimation. In the second stage, the pairwise interaction branch and post-$\tau$ fusion module are activated, still with the encoder frozen, so that collision relevant refinement can be learned on top of an already stable geometric prior. In the final stage, all parameters are jointly fine-tuned, enabling scene encoding, interaction reasoning, and timing adjustment to be co-adapted.

The contribution of different supervision signals is not kept constant throughout optimization. Instead, stage loss warmup is applied so that collision related, semantic, and behavior-aware losses are introduced progressively rather than dominating early training. In parallel, the teacher-forcing ratio associated with the transition variable $\tau$ is gradually reduced across stages, allowing the model to move smoothly from strongly guided temporal supervision toward autonomous transition-time inference.

\subsection{Evaluation results}

\begin{figure*}[t]
\centering 
\includegraphics[width=0.95\textwidth]{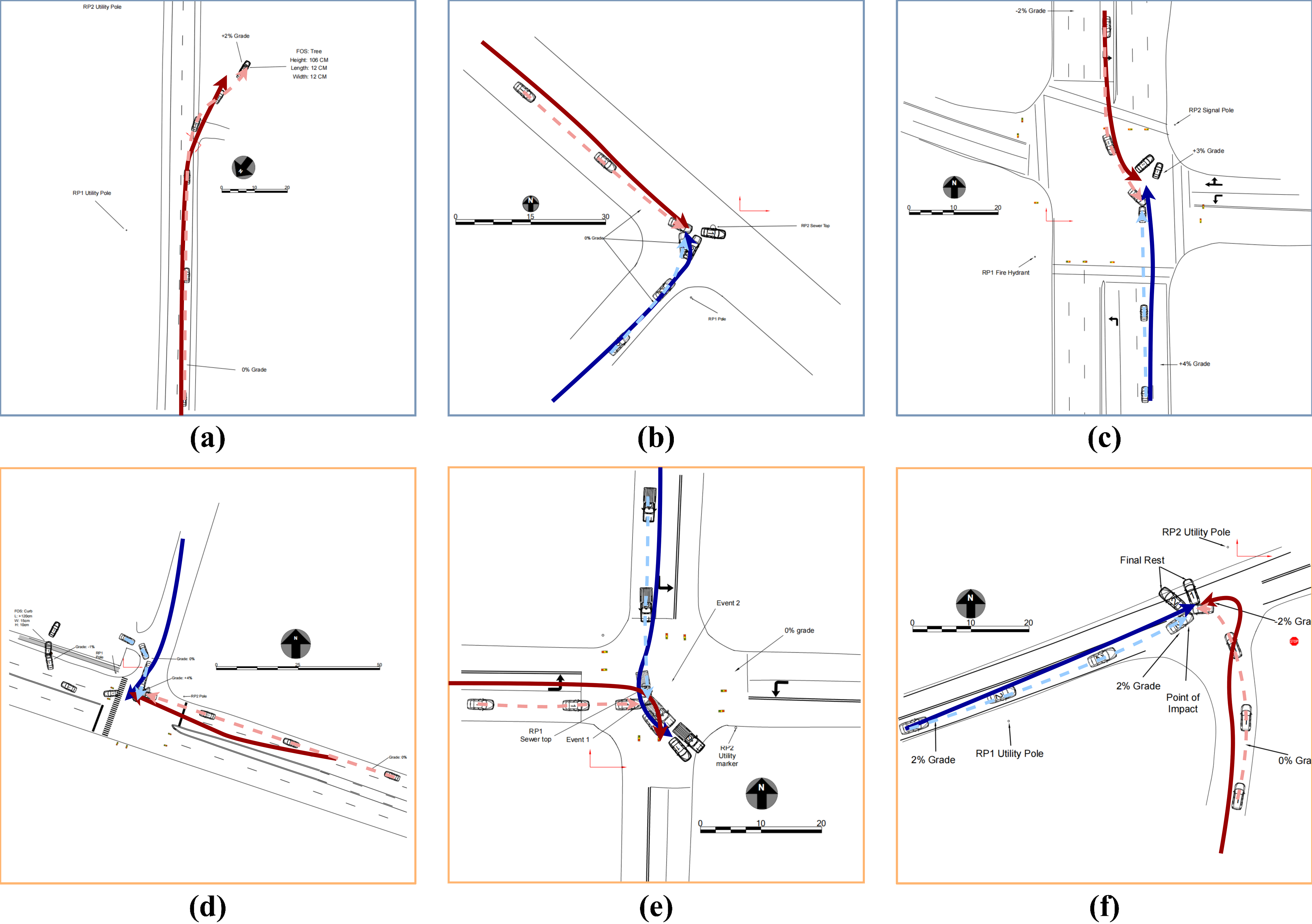}
\caption{Visualization results of representative accident cases. The dark solid lines denote the inferred trajectories generated by the proposed model, while the light dashed lines indicate the ground-truth trajectories recorded during the actual traffic accidents.}
\label{result_vis}
\end{figure*}

\begin{figure*}[t]
\centering 
\includegraphics[width=0.95\textwidth]{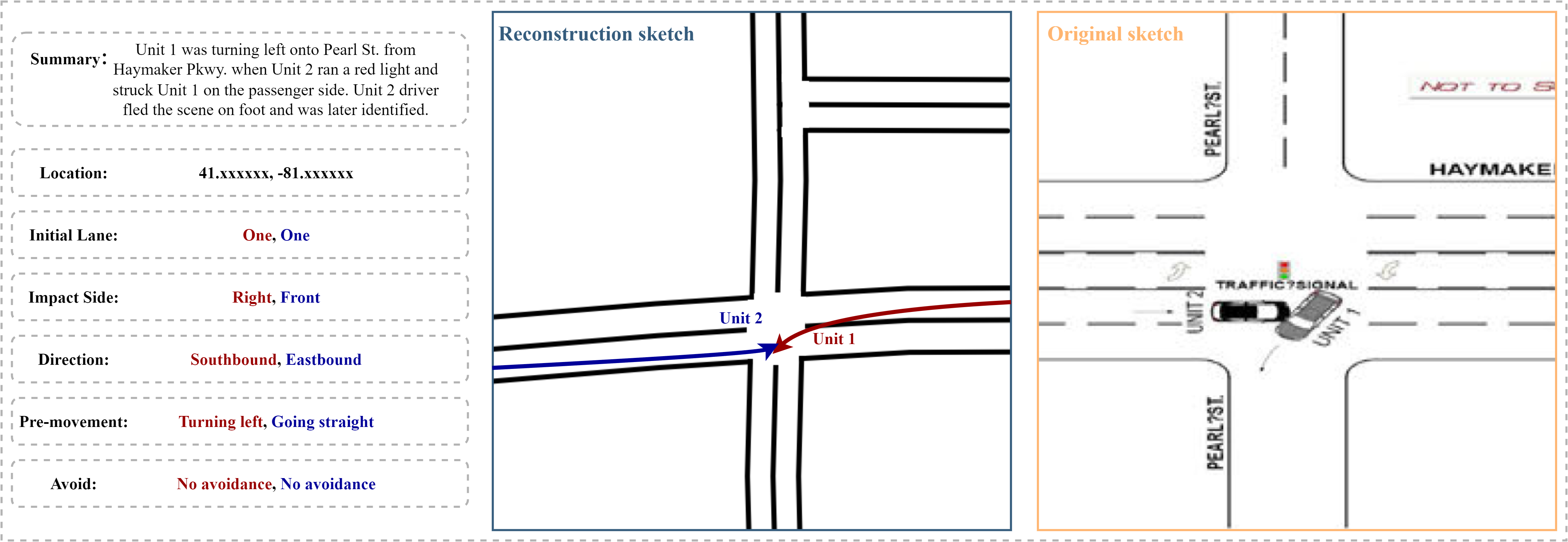}
\caption{Visualization of an external traffic accident report. Some of the input information is inferred from the original accident report using Qwen3. To avoid privacy breaches, we have blurred the specific coordinates.}
\label{result_vis2}
\end{figure*}

Table~\ref{results} presents the experimental results of our proposed framework and several baseline models for traffic accident trajectory reconstruction. To better simulate the real-world scenario of non-deep accident investigation, we injected random Gaussian noise with a maximum value of 1~m into the accident site before conducting the experiment to simulate real-world positioning errors.
The selected baselines are broadly representative of mainstream trajectory modeling paradigms, including recurrent sequence models, graph-based interaction models, vectorized map encoders, multimodal forecasting frameworks, and transformer-based architectures. All models use the same input, training process, and supervision. It is worth noting that PC-Crash and the Momentum-Energy, as classical baselines specifically designed for collision reconstruction, explicitly force the involved vehicles to converge to the same spatial point to form a collision. This design naturally leads to extremely high collision rates for both methods. However, such forced alignment does not preserve realistic interaction dynamics, resulting in unsatisfactory performance in terms of collision surface accuracy, scene-level semantic consistency, and other dynamical metrics.

As described in Section~\ref{Evaluation metrics}, the evaluation metrics serve different roles in assessing reconstruction quality. Specifically, AKD mainly measures the discrepancy between the overall reconstructed trajectories and the ground-truth trajectories, reflecting global spatial accuracy. AVD evaluates the reconstruction quality of vehicle speed and characterizes the consistency of motion dynamics. In contrast, AAPD, CR, and CSA focus on the reconstruction quality near the collision phase, measuring the accuracy of collision localization, collision occurrence, and collision contact geometry, respectively. In addition, to measure the physical consistency between the reconstructed vehicle motion and the real accident data, we compared the average acceleration error (Acc) and curvature error (Curvature) of the vehicle trajectories reconstructed by different models.

Figure~\ref{result_hot} presents a heatmap constructed from the experimental results, providing an intuitive visualization of the performance of different baseline models across various evaluation metrics. The experimental results show that the proposed model significantly outperforms the general accident reconstruction baselines in terms of both spatial and temporal consistency of reconstructed accident trajectories. In particular, the proposed method yields the lowest spatial and collision point errors ($AKD$,$AAPD_{tan \& norm}$), while substantially improving collision recovery and semantic consistency ($CR$, $CSA$, and $BA$). These results indicate that the model does not merely generate plausible trajectories, but more effectively reconstructs the coupled geometric, dynamic and semantic structure of real-world accident processes. This improvement stems from our semantic--geometric--dynamical coupled conditional inverse inference framework, which enables the model to maintain strong spatial alignment with the underlying road topology and collision regions while simultaneously recovering realistic dynamic patterns of accident evolution, including braking behavior, evasive maneuvers, and the temporal progression leading to near-collision and impact events.  

However, the velocity reconstruction error in our framework remains relatively large. This limitation primarily arises from the fact that accident reports typically provide only the posted speed limit of the roadway, rather than the actual driving speeds of individual vehicles. As a result, the model must infer plausible velocity profiles under weak supervision, which makes learning accurate speed patterns more challenging, particularly when trajectory reconstruction and dynamic consistency must be optimized simultaneously.
Despite this limitation, the superior performance of our framework in behavior accuracy indicates that, although the absolute velocity error is slightly higher than that of some baseline models, our method better captures the overall motion patterns that jointly characterize vehicle trajectories and speed evolution. In other words, the reconstructed trajectories and velocity profiles remain dynamically consistent with the behavioral semantics of the accident process.
To further investigate this issue, we conducted an additional experiment in which the maximum velocity recorded in the EDR data was used to replace the roadway speed limit as the velocity constraint. Under this setting, the $AVD$ decreased to $2.55$, providing additional evidence that the primary source of velocity error originates from the limited speed information available in accident reports rather than from deficiencies in the proposed reconstruction framework.

In terms of acceleration and curvature, the proposed model also significantly outperforms other baselines, indicating that the reconstructed vehicle motions better adhere to real-world physical dynamics in both velocity evolution and trajectory variation.
Notably, simple kinematic fitting also achieves competitive performance. Despite performing substantially worse in most other metrics, it even yields lower curvature error than the proposed model. This observation suggests that real-world vehicle motion tends to exhibit relatively smooth and stable patterns, which can be partially captured by simplified kinematic assumptions.

Figure~\ref{result_vis} presents visualization results of several representative cases. In the scenarios shown in Fig.~\ref{result_vis}(a)-(c), the proposed model successfully reconstructs the accident scenes with relatively small errors, and the inferred trajectories closely match the ground-truth vehicle motions.

On the contrary, in the cases shown in Fig.~\ref{result_vis}(d)-(f), noticeable deviations exist between the inferred trajectories and the actual accident trajectories. In Fig.~\ref{result_vis}(d), the map corresponding to the lane from which the blue vehicle enters the scene is incomplete. As a result, the model selects an incorrect candidate lane during the lane assignment stage, leading to a significant shift in the initial position of the blue vehicle. This initial misalignment propagates through the trajectory inference process, producing a large overall deviation from the real accident trajectory.

In the case shown in Fig.~\ref{result_vis}(e), although the two vehicles successfully collide, the overall scenario still deviates from the original accident, particularly when considering the vehicle speed and position at specific time steps. This issue mainly arises because the input map emphasizes geometric shape information rather than topological constraints. Consequently, the model fails to fully capture lane-level turning restrictions, allowing vehicles to proceed straight or turn right from lanes that are intended for left-turn movements only. Although the resulting deviation corresponds to a shift of approximately one lane, this discrepancy leads to a substantial difference between the reconstructed scenario and the actual accident configuration.

Furthermore, the case shown in Fig.~\ref{result_vis}(f) reveals another important source of reconstruction error, namely the insufficient granularity of collision surface annotations. In the real accident, the two vehicles collide head-on at a certain angle; however, this angular information is not explicitly captured in the structured data. As a result, during reconstruction, the model tends to enforce a strictly frontal collision.
To satisfy this constraint while maintaining physical plausibility and accurate collision positioning, the model reconstructs a large arc trajectory for the red vehicle prior to impact. This behavior deviates significantly from realistic driving patterns and human driving habits, ultimately leading to noticeable discrepancies in the reconstructed scene.

To further evaluate the generalization capability of the proposed model, Figure~\ref{result_vis2} presents a reconstruction result based on a traffic accident report obtained from the Ohio Department of Public Safety. In this experiment, key structured inputs, including \emph{Initial Lane}, \emph{Pre-movement}, and \emph{Avoidance behavior}, are extracted from the accident description using Qwen3. A high-resolution map of the area near the accident point, downloaded from OpenStreetMap, is used as the input map.

Although quantitative evaluation is not available due to the lack of parameterized ground-truth annotations in this data source, qualitative analysis shows that the reconstructed trajectories are largely consistent with the semantic content of the accident report. In particular, the reconstructed scene preserves the key interaction patterns and event structure described in the original report, indicating that the model is capable of generalizing to unseen data sources and recovering semantically coherent accident scenarios.

However, during this process we also identified an important limitation. When using real-world high-definition maps, inferring the initial vehicle position based solely on the initial lane and direction can be ambiguous. Specifically, when multiple parallel roads share the same direction, the model cannot reliably determine which road the vehicle actually belongs to. 

In contrast, this issue does not arise when using sketch-based inputs, since such sketches typically include only the road structures directly relevant to the accident, thereby eliminating ambiguity. This observation highlights that, in real-world traffic accident reconstruction, recording additional contextual information, such as the exact road name, is crucial for resolving spatial ambiguity and improving reconstruction accuracy.

\subsection{Ablation studies}

\begin{table*}[t]
\centering
\caption{Ablation study of key modules of our proposed model. Distance-based metrics are reported in $m$, velocity-based metrics in $m/s$, and accuracy-based metrics in \%.}
 % \resizebox{0.8\linewidth}{!}{
\begin{tabular}{c c c cc c c }
\hline \specialrule{0em}{1pt}{1pt}
\multirow{2}{*}{Model} & \multirow{2}{*}{AKD} & \multirow{2}{*}{AVD} & \multicolumn{2}{c}{AAPD} & \multirow{2}{*}{CR} & \multirow{2}{*}{CSA} \\
\cline{4-5}
& & & tan & norm & & \\
\hline
w/o Post-$\tau$ fusion & 7.44 & 8.39 & 8.04  & 3.82 & 27.27 & 20.45 \\
w/o Graph transformer decoder & 7.21 & 7.25 & 7.69  & 3.55 & 25.76 & 21.97 \\
w/o Lane candidate& 6.40 & 7.53 & 5.73  & 2.81 & 86.36 & 41.67 \\
w/o Stage-wise training & 6.25 & 8.11 &  3.09 & 1.20 & 86.36 & 58.21 \\
w/o Temporal allocation  & 4.45 & 9.44 & 3.10  & 1.15 & 85.45 & 53.03 \\
w/o EDR supervision & 4.64 & 9.35 & 3.09  & 1.21 & 86.25 & 51.34 \\ 

\hline
Full &4.98 & 6.84 & 3.47 & 1.41 & 85.71 & 56.35 \\ \hline
\end{tabular}
% }

\label{AblationMODULE}
\end{table*}

\begin{table*}[t]
\centering
\caption{Ablation study of input attributes in accident reports. Distance-based metrics are reported in $m$, velocity-based metrics in $m/s$, and accuracy-based metrics in \%.}
 % \resizebox{0.8\linewidth}{!}{
\begin{tabular}{c c c cc c c }
\hline \specialrule{0em}{1pt}{1pt}
\multirow{2}{*}{Model} & \multirow{2}{*}{AKD} & \multirow{2}{*}{AVD} & \multicolumn{2}{c}{AAPD} & \multirow{2}{*}{CR} & \multirow{2}{*}{CSA} \\
\cline{4-5}
& & & tan & norm & & \\
\hline
w/o Speed limit & 24.60 & 13.14 & 11.39  & 4.64 & 81.82 & 45.52 \\
w/o Initial line & 16.75 & 8.24 & 5.96  & 5.18 &  68.18 & 44.78 \\
w/o Location & 9.83 & 8.25  & 11.74 & 5.85 & 90.91  & 54.48\\ 
w/o Avoid & 6.45 & 7.35 &  3.31 & 1.85 & 83.33 & 50.75 \\
w/o Pre-movement & 5.81 & 7.51  &  2.86 & 1.49 & 87.88 & 55.30 \\
w/o Summary & 5.67 & 7.78 &  3.59 & 1.46 & 86.36 & 56.82 \\
w/o Impact side & 5.50 & 8.64  &  2.96 & 1.53 & 85.91 & 34.09 \\
w/o Direction & 5.26  & 8.28 & 3.46 & 1.41 & 89.39  & 48.51 \\ 
\hline
Full &4.98 & 6.84 & 3.47 & 1.41 & 85.71 & 56.35 \\ \hline
\end{tabular}
% }

\label{AblationINPUT}
\end{table*}

\begin{table*}[t]
\centering
\caption{Experimental results on the robustness of the model against specific input noise. Distance-based metrics are reported in $m$, velocity-based metrics in $m/s$, and accuracy-based metrics in \%.}
 % \resizebox{0.8\linewidth}{!}{
\begin{tabular}{c c c cc c c }
\hline \specialrule{0em}{1pt}{1pt}
\multirow{2}{*}{Eval} & \multirow{2}{*}{AKD} & \multirow{2}{*}{AVD} & \multicolumn{2}{c}{AAPD} & \multirow{2}{*}{CR} & \multirow{2}{*}{CSA} \\
\cline{4-5}
& & & tan & norm & & \\
\hline
w/o Speed limit & 7.08 & 7.63 & 3.47  & 1.35 & 84.04 & 61.73 \\
w/o Initial line & 13.59 & 6.81 & 1.96  & 1.23 &  88.32 & 47.52 \\
w/o Location & 5.68 & 6.93  & 4.28 & 3.06 & 85.10  & 56.79\\ 
w/o Avoid & 4.68 & 7.14 &  3.81 & 1.58 & 85.82 & 56.36 \\
w/o Pre-movement & 5.99 & 7.49  &  2.80 & 1.22 & 88.67 & 59.22 \\
w/o Summary & 4.93 & 6.93 &  3.45 & 1.45 & 84.35 & 57.06 \\
w/o Impact side & 4.95 & 6.92  &  3.52 & 1.48 & 86.35 & 40.68 \\
w/o Direction & 5.18  & 6.90 & 3.72 & 1.57 & 84.75  & 51.42 \\ 
60\% map & 6.83  & 6.86 & 3.29 & 1.43 & 86.89  & 48.56\\ 

\hline
Full &4.98 & 6.84 & 3.47 & 1.41 & 85.71 & 56.35 \\ \hline
\end{tabular}
% }

\label{RobustINPUT}
\end{table*}

\begin{table*}[t]
\centering
\caption{Experimental results on the robustness of the model against random entry missing. Distance-based metrics are reported in $m$, velocity-based metrics in $m/s$, and accuracy-based metrics in \%.}
 % \resizebox{0.8\linewidth}{!}{
\begin{tabular}{c c c cc c c }
\hline \specialrule{0em}{1pt}{1pt}
\multirow{2}{*}{Eval} & \multirow{2}{*}{AKD} & \multirow{2}{*}{AVD} & \multicolumn{2}{c}{AAPD} & \multirow{2}{*}{CR} & \multirow{2}{*}{CSA} \\
\cline{4-5}
& & & tan & norm & & \\
\hline
3\% entry missing & 5.16  & 7.05 & 3.46 & 1.46 & 85.46  & 55.19\\
5\% entry missing & 5.32  & 7.03 & 3.41 & 1.46 & 85.28  & 53.51\\
10\% entry missing & 6.08  & 7.10 & 3.36 & 1.47 & 84.39  & 51.09\\
15\% entry missing & 6.55  & 7.23 & 3.27 & 1.43 & 84.39  & 49.26\\
20\% entry missing & 7.27  & 7.35 & 3.19 & 1.44 & 83.85  & 47.55\\
30\% entry missing & 8.78 & 7.46 & 3.09 & 1.43 & 83.67  & 43.33\\
50\% entry missing & 11.59  & 7.81 & 2.92 & 1.43 & 84.75  & 38.44\\
\hline
Full &4.98 & 6.84 & 3.47 & 1.41 & 85.71 & 56.35 \\ \hline
\end{tabular}
% }

\label{RobustEntry}
\end{table*}

\begin{figure*}[t]
\centering 
\includegraphics[width=0.90\textwidth]{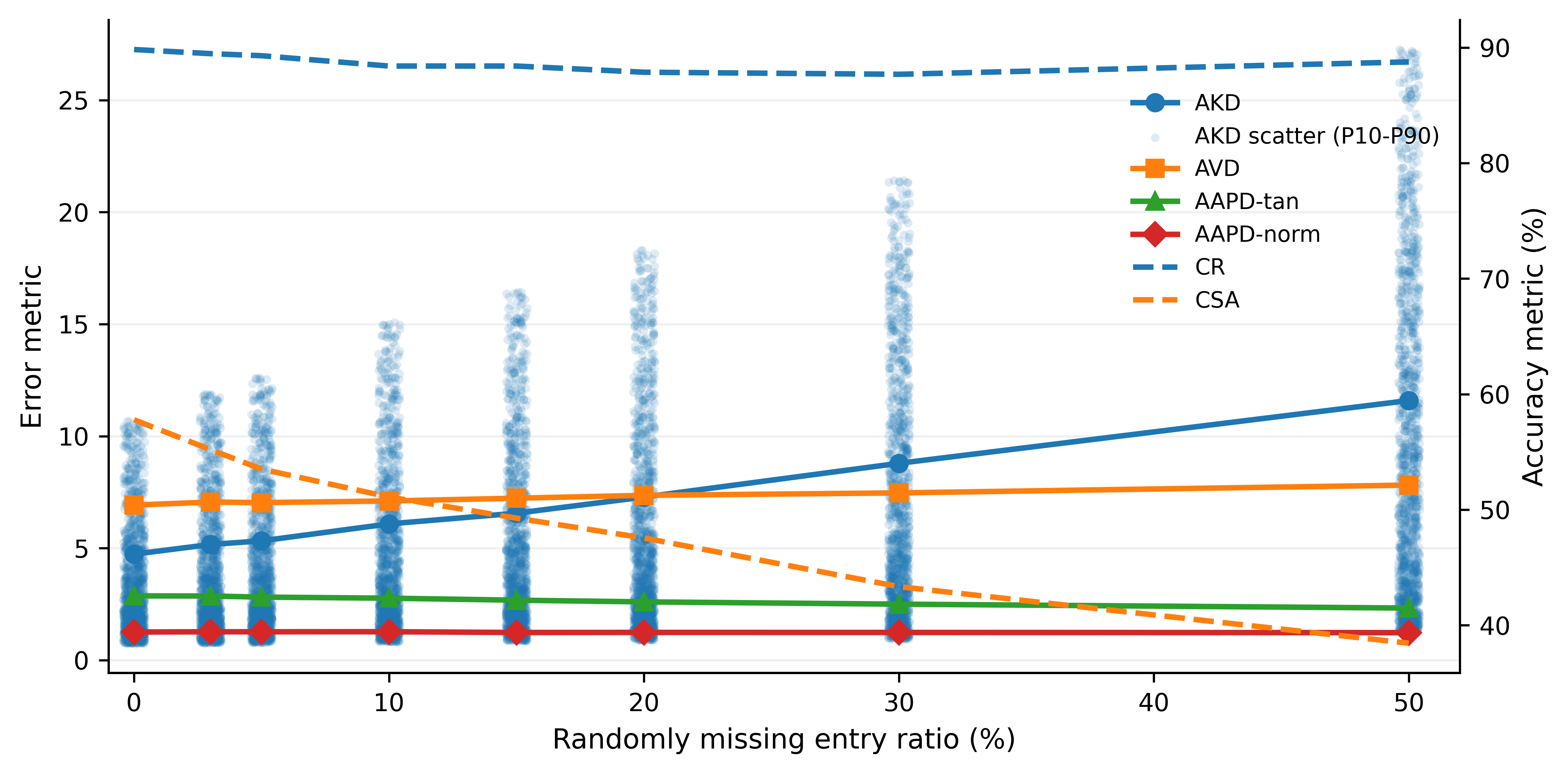}
\caption{Sensitivity of reconstruction metrics to random report-entry missingness.}
\label{missing}
\end{figure*}

Table~\ref{AblationMODULE} presents the ablation study conducted on the CISS-REC dataset, evaluating the influence of key modules within the proposed framework on traffic accident trajectory reconstruction performance. Compared with the balanced reconstruction capability of the full model, the effect of \emph{Stage-wise training} mainly lies in improving training stability and facilitating coordinated optimization among modules. By progressively introducing different components during training, the framework allows each submodule to establish cooperative relationships in a more structured manner, leading to a more stable learning process. After ablating this module, both AKD and AVD increase significantly, indicating that jointly learning trajectory and velocity without staged optimization introduces strong optimization conflicts, preventing the model from converging to a physically consistent solution.

The \emph{Lane candidate} module serves as one of the core sources of geometric-semantic priors. It assists the model in determining the roadway affiliation, driving corridor, and geometric contact relationships of vehicles prior to the accident. Consequently, this module plays a crucial role in accurately localizing accident points and recovering collision contact surfaces. Due to trajectory continuity, the initial lane selection globally influences the entire trajectory, including both the initial position and subsequent motion evolution. Therefore, ablating this module leads to performance degradation across nearly all metrics.

The \emph{Temporal allocation} module primarily focuses on velocity modeling and temporal coordination. It enables the model to learn the temporal organization of trajectory evolution, including speed distribution, motion rhythm, and the timing structure leading to collision events. After removing this module, the inferred velocity error increases substantially, while trajectory-related errors slightly decrease. This observation further confirms the optimization conflict between trajectory and velocity learning during joint training.

Finally, the \emph{Post-$\tau$ fusion} and \emph{Graph transformer decoder} constitute the backbone modules responsible for collision reconstruction capability. Without explicit mechanisms for modeling inter-vehicle interactions and learning the temporal boundary between normal driving and evasive maneuvers, the model would struggle to capture the most fundamental aspect of traffic accidents—namely, trajectory deflections, contact formation, and geometric collision evolution resulting from vehicle interactions. Ablating these two modules not only degrades overall trajectory and velocity inference performance, but also directly affects trajectory learning near the collision moment, making it difficult for the reconstructed vehicles to collide in accordance with the accident description.

In addition, to investigate how the content of traffic accident reports influences reconstruction performance, we conduct an input ablation study by selectively removing specific report attributes from the model input. The results are summarized in Table~\ref{AblationINPUT}. The experiments indicate that structured factors directly constraining vehicle dynamics boundaries, roadway affiliation, and collision contact relationships play a decisive role in reconstruction accuracy.

The \emph{Speed limit}, as the only speed-related information available in accident reports, acts as a crucial physical boundary for the entire accident dynamics. It not only affects velocity prediction itself but also participates deeply in modeling trajectory length, collision arrival time, and the spatial scale of motion evolution. The \emph{Initial lane} directly determines the roadway affiliation of the vehicle prior to the accident, its entry path, and the relative interaction positions among participants, making it closely tied to spatial trajectory reconstruction. In contrast, the \emph{Location} attribute mainly determines the approximate spatial region where the accident occurs. Without location information, the model is more likely to achieve a higher CR through coarse collision matching strategies, but this substantially degrades spatial realism and scene consistency.

For other semantic attributes, ablating a single factor results in relatively limited performance degradation. This is because many of these attributes serve as auxiliary conditions and partially overlap with each other in semantic content. For example, directional information and driving behaviors are often implicitly contained within the overall accident description. This observation is also consistent with the structure of real-world accident reports, where some reports do not require an explicit narrative summary but instead record accident details through a set of structured fields.

Notably, removing the \emph{Impact side} attribute leads to the largest decrease in CSA, highlighting its critical role in recovering the contact geometry of collisions rather than affecting the global trajectory reconstruction quality.

\subsection{Robustness studies}
To evaluate the robustness of the proposed model under potential data noise, we keep the testing checkpoint fixed and measure performance under different input-missing conditions, as shown in Table~\ref{RobustINPUT}. For the model trained with complete inputs, as discussed previously, semantic redundancy and overlap among textual attributes reduce sensitivity to partial missing information. Consequently, removing individual textual fields produces only marginal changes in overall performance.

In contrast, the removal of \emph{Initial lane}, which determines the initial vehicle placement, and \emph{Location}, which constrains the spatial range of the accident, has a clear impact on both AKD and AAPD. These two attributes directly affect the global spatial configuration of the reconstructed trajectories, making them critical for accurate accident reconstruction.

The absence of \emph{Speed limit} affects both trajectory length and velocity inference, since it acts as a physical constraint on vehicle dynamics. As a result, removing this attribute leads to increases in both AKD and AVD, reflecting degraded spatial and dynamical consistency.

Furthermore, we evaluate the impact of incomplete \emph{map geometry} by randomly removing $40\%$ of map polylines. Under this setting, AKD increases noticeably, indicating that the overall trajectory reconstruction is significantly affected. However, the influence on velocity estimation and accident-point localization remains relatively small, suggesting that map geometry primarily affects spatial trajectory alignment rather than temporal dynamics or collision positioning.

In addition to robustness evaluation under specific missing attributes, we further conduct random ablation of textual inputs at different proportions to simulate the effect of stochastic input incompleteness. The results are summarized in Table~\ref{RobustEntry}. As the proportion of missing text increases, both AKD and AVD (metrics associated with global trajectory reconstruction) gradually increase, indicating that the model becomes less capable of accurately localizing the overall trajectory under insufficient information.

In contrast, due to the presence of the Post-$\tau$ fusion module, the collision rate (CR) remains relatively stable and fluctuates within a normal range, showing limited sensitivity to missing textual inputs. Moreover, as the model struggles to learn the overall trajectory under increasing information loss, the optimization process becomes more focused on the collision region, which in turn leads to improved performance in AAPD.

Finally, as the amount of missing collision-related semantic information increases, CSA gradually deteriorates. This trend is consistent with the expectation that collision surface accuracy is highly dependent on detailed semantic cues describing contact geometry.

Figure~\ref{missing} clearly illustrates the process of structural degradation as the proportion of missing textual inputs increases. In particular, both AKD and CSA exhibit an approximately linear degradation trend when the missing rate remains below $50\%$, indicating a steady decline in reconstruction quality due to the loss of semantic information.
Furthermore, we analyze the distribution of AKD at the per-vehicle level. The results show that the variance of AKD gradually increases as more textual information is removed, leading to a wider error distribution. This observation highlights the critical role of original textual inputs in constraining the global trajectory reconstruction of individual vehicles.

\section{Conclusion}

In this work, we formulated traffic accident reconstruction from publicly accessible accident reports and scene measurements as a parameterized multimodal learning problem. To support this task, we constructed CISS-REC, a real-world accident reconstruction dataset curated from the NHTSA Crash Investigation Sampling System, and developed a scene-grounded reconstruction framework that integrates report semantics, road geometry, and sparse motion evidence into a unified reconstruction process. 

Experimental results show that the proposed framework achieves the strongest overall reconstruction fidelity among representative baselines, with clear advantages in accident-point localization and collision consistency. These findings suggest that publicly available accident reports can support quantitatively verifiable accident reconstruction beyond coarse semantic restoration. We further demonstrated that the reconstructed trajectories can be coupled with a downstream controllable rendering pipeline, enabling visually interpretable multi-view replay and post-impact scene extension.

Overall, this study provides a scalable route from public accident reports to scene-specific accident reconstruction, with potential value for transportation safety analysis, simulation-based investigation, and autonomous-driving research.

\section*{Acknowledgment}

This work was supported by the Science and Technology Development Fund of Macau [0007/2025/RIC, 0122/2024/RIB2, 0215/2024/AGJ, 0074/2025/AMJ, 001/2024/SKL, 0002/2025/EQP], the Research Services and Knowledge Transfer Office, University of Macau [SRG2023-00037-IOTSC, MYRG-GRG2024-00284-IOTSC], the Shenzhen-Hong Kong-Macau Science and Technology Program Category C [SGDX20230821095159012], the Science and Technology Planning Project of Guangdong [2025A0505010016], National Natural Science Foundation of China [52572354], the State Key Lab of Intelligent Transportation System [2024-B001], and the Jiangsu Provincial Science and Technology Program [BZ2024055].

\section*{Data availability}
The data used in this work will be publicly released.

\appendix

% \printcredits

%% Loading bibliography style file
\bibliographystyle{model1-num-names}

\bibliography{cas-refs}

@article{bertolazzi2018g2,
  title={On the G2 Hermite interpolation problem with clothoids},
  author={Bertolazzi, Enrico and Frego, Marco},
  journal={Journal of Computational and Applied Mathematics},
  volume={341},
  pages={99--116},
  year={2018},
  publisher={Elsevier}
}

@article{griffin2020automatic,
  title={Automatic collision notification availability and emergency response times following vehicle collision—An analysis of the 2017 crash investigation sampling system},
  author={Griffin, Russell L and Carroll, Shannon and Jansen, Jan O},
  journal={Traffic injury prevention},
  volume={21},
  number={sup1},
  pages={S135--S139},
  year={2020},
  publisher={Taylor \& Francis}
}

@inproceedings{tan2019text2scene,
  title={Text2scene: Generating compositional scenes from textual descriptions},
  author={Tan, Fuwen and Feng, Song and Ordonez, Vicente},
  booktitle={Proceedings of the IEEE/CVF Conference on Computer Vision and Pattern Recognition},
  pages={6710--6719},
  year={2019}
}

@article{wach2016calculation,
  title={Calculation reliability in vehicle accident reconstruction},
  author={Wach, Wojciech},
  journal={Forensic science international},
  volume={263},
  pages={27--38},
  year={2016},
  publisher={Elsevier}
}

@article{beck2023automated,
  title={Automated vehicle data pipeline for accident reconstruction: New insights from LiDAR, camera, and radar data},
  author={Beck, Joe and Arvin, Ramin and Lee, Steve and Khattak, Asad and Chakraborty, Subhadeep},
  journal={Accident Analysis \& Prevention},
  volume={180},
  pages={106923},
  year={2023},
  publisher={Elsevier}
}

@article{lee2023advancing,
  title={Advancing investigation of automated vehicle crashes using text analytics of crash narratives and Bayesian analysis},
  author={Lee, Steve and Arvin, Ramin and Khattak, Asad J},
  journal={Accident Analysis \& Prevention},
  volume={181},
  pages={106932},
  year={2023},
  publisher={Elsevier}
}

@article{arteaga2020injury,
  title={Injury severity on traffic crashes: A text mining with an interpretable machine-learning approach},
  author={Arteaga, Cristian and Paz, Alexander and Park, JeeWoong},
  journal={Safety Science},
  volume={132},
  pages={104988},
  year={2020},
  publisher={Elsevier}
}

@article{savolainen2011statistical,
  title={The statistical analysis of highway crash-injury severities: A review and assessment of methodological alternatives},
  author={Savolainen, Peter T and Mannering, Fred L and Lord, Dominique and Quddus, Mohammed A},
  journal={Accident Analysis \& Prevention},
  volume={43},
  number={5},
  pages={1666--1676},
  year={2011},
  publisher={Elsevier}
}

@article{lord2010statistical,
  title={The statistical analysis of crash-frequency data: A review and assessment of methodological alternatives},
  author={Lord, Dominique and Mannering, Fred},
  journal={Transportation research part A: policy and practice},
  volume={44},
  number={5},
  pages={291--305},
  year={2010},
  publisher={Elsevier}
}

@article{liu2023integrated,
  title={An integrated data-and theory-driven crash severity model},
  author={Liu, Dongjie and Li, Dawei and Sze, NN and Ding, Hongliang and Song, Yuchen},
  journal={Accident Analysis \& Prevention},
  volume={193},
  pages={107282},
  year={2023},
  publisher={Elsevier}
}

@article{li2024steering,
  title={Steering the future: Redefining intelligent transportation systems with foundation models},
  author={Li, Zhenning and Cui, Zhiyong and Liao, Haicheng and Ash, John and Zhang, Guohui and Xu, Chengzhong and Wang, Yinhai},
  journal={Chain},
  volume={1},
  number={1},
  pages={46--53},
  year={2024},
  publisher={Youke Publishing}
}

@inproceedings{guo2024sovar,
  title={Sovar: Build generalizable scenarios from accident reports for autonomous driving testing},
  author={Guo, An and Zhou, Yuan and Tian, Haoxiang and Fang, Chunrong and Sun, Yunjian and Sun, Weisong and Gao, Xinyu and Luu, Anh Tuan and Liu, Yang and Chen, Zhenyu},
  booktitle={Proceedings of the 39th IEEE/ACM International Conference on Automated Software Engineering},
  pages={268--280},
  year={2024}
}

@article{li2025avd2,
  title={Avd2: Accident video diffusion for accident video description},
  author={Li, Cheng and Zhou, Keyuan and Liu, Tong and Wang, Yu and Zhuang, Mingqiao and Gao, Huan-ang and Jin, Bu and Zhao, Hao},
  journal={arXiv preprint arXiv:2502.14801},
  year={2025}
}

@article{zhang2025accidentsim,
  title={Accidentsim: Generating physically realistic vehicle collision videos from real-world accident reports},
  author={Zhang, Xiangwen and Zhang, Qian and Han, Longfei and Qu, Qiang and Chen, Xiaoming},
  journal={arXiv preprint arXiv:2503.20654},
  year={2025}
}

@article{li2025crashagent,
  title={CrashAgent: Crash Scenario Generation via Multi-modal Reasoning},
  author={Li, Miao and Ding, Wenhao and Lin, Haohong and Lyu, Yiqi and Yao, Yihang and Zhang, Yuyou and Zhao, Ding},
  journal={arXiv preprint arXiv:2505.18341},
  year={2025}
}

@article{zheng2020determinants,
  title={Determinants of the congestion caused by a traffic accident in urban road networks},
  author={Zheng, Zhenjie and Wang, Zhengli and Zhu, Liyun and Jiang, Hai},
  journal={Accident Analysis \& Prevention},
  volume={136},
  pages={105327},
  year={2020},
  publisher={Elsevier}
}

@book{struble2020automotive,
  title={Automotive accident reconstruction: practices and principles},
  author={Struble, Donald E and Struble, John D},
  year={2020},
  publisher={CRC Press}
}

@incollection{fernandes2018application,
  title={Application of numerical methods for accident reconstruction and forensic analysis},
  author={Fernandes, F{\'a}bio AO and Alves de Sousa, Ricardo J and Ptak, Mariusz},
  booktitle={Head Injury Simulation in Road Traffic Accidents},
  pages={59--98},
  year={2018},
  publisher={Springer}
}

@incollection{ryan2024accident,
  title={Accident Investigation Processes and Techniques in Sociotechnical Systems},
  author={Ryan, David S and Zarei, Esmaeil},
  booktitle={Safety Causation Analysis in Sociotechnical Systems: Advanced Models and Techniques},
  pages={21--45},
  year={2024},
  publisher={Springer}
}

@book{rivers2010technical,
  title={Technical Traffic Crash Investigators' Handbook:(level 3): a Technical Reference, Training, Investigation and Reconstruction Manual},
  author={Rivers, Robert W},
  year={2010},
  publisher={Charles C Thomas Publisher}
}

@book{rivers2006evidence,
  title={Evidence in traffic crash investigation and reconstruction: identification, interpretation and analysis of evidence, and the traffic crash investigation and reconstruction process},
  author={Rivers, Robert W},
  year={2006},
  publisher={Charles C Thomas Publisher}
}

@article{mohammed2023overview,
  title={An overview of traffic accident investigation using different techniques},
  author={Mohammed, Shireen Ibrahim},
  journal={Automotive experiences},
  volume={6},
  number={1},
  pages={68--79},
  year={2023}
}

@article{vida2023analysis,
  title={Analysis of UAV flight patterns for road accident site investigation},
  author={Vida, G{\'a}bor and Melegh, G{\'a}bor and S{\"u}veges, {\'A}rp{\'a}d and Wenszky, N{\'o}ra and T{\"o}r{\"o}k, {\'A}rp{\'a}d},
  journal={Vehicles},
  volume={5},
  number={4},
  pages={1707--1726},
  year={2023},
  publisher={MDPI}
}

@article{su2016developing,
  title={Developing an unmanned aerial vehicle-based rapid mapping system for traffic accident investigation},
  author={Su, Sen and Liu, Wenjun and Li, Kui and Yang, Guangyu and Feng, Chengjian and Ming, Jianxiong and Liu, Guodong and Liu, Shengxiong and Yin, Zhiyong},
  journal={Australian journal of forensic sciences},
  volume={48},
  number={4},
  pages={454--468},
  year={2016},
  publisher={Taylor \& Francis}
}

@article{jiang2021unmanned,
  title={Unmanned Aerial Vehicle-Based Photogrammetric 3D Mapping: A survey of techniques, applications, and challenges},
  author={Jiang, San and Jiang, Wanshou and Wang, Lizhe},
  journal={IEEE Geoscience and Remote Sensing Magazine},
  volume={10},
  number={2},
  pages={135--171},
  year={2021},
  publisher={IEEE}
}

@incollection{lemmens2011terrestrial,
  title={Terrestrial laser scanning},
  author={Lemmens, Mathias},
  booktitle={Geo-information: technologies, applications and the environment},
  pages={101--121},
  year={2011},
  publisher={Springer}
}

@article{scherer2009conventional,
  title={From the conventional total station to the prospective image assisted photogrammetric scanning total station: Comprehensive review},
  author={Scherer, Michael and Lerma, Jos{\'e} Luis},
  journal={Journal of Surveying Engineering},
  volume={135},
  number={4},
  pages={173--178},
  year={2009},
  publisher={American Society of Civil Engineers}
}

@techreport{clamann2021advancing,
  title={Advancing crash investigation with connected and automated vehicle data},
  author={Clamann, Michael and Khattak, Asad J and Clark, Kinzee and others},
  year={2021},
  institution={Collaborative Sciences Center for Road Safety}
}

@inproceedings{dhanam2025event,
  title={Event Data Recorder for Investigation, Legal Claim and Fault Analysis in Vehicles},
  author={Dhanam, B and Marichamy, P and Mohan, B and Manoj, E and Johnson, J and Balavignesh, K},
  booktitle={2025 3rd International Conference on Artificial Intelligence and Machine Learning Applications Theme: Healthcare and Internet of Things (AIMLA)},
  pages={1--5},
  year={2025},
  organization={IEEE}
}

@article{raviv2017analyzing,
  title={Analyzing risk factors in crane-related near-miss and accident reports},
  author={Raviv, Gabriel and Fishbain, Barak and Shapira, Aviad},
  journal={Safety science},
  volume={91},
  pages={192--205},
  year={2017},
  publisher={Elsevier}
}

@book{carper2000forensic,
  title={Forensic engineering},
  author={Carper, Kenneth L},
  year={2000},
  publisher={CRC press}
}

@article{faizan2021forensic,
  title={Forensic investigation of road traffic accident cases in Pakistan and Types of Physical Evidence},
  author={Faizan, Khurram and Abid, Adeel},
  journal={Pakistan Social Sciences Review},
  volume={5},
  number={4},
  pages={405--422},
  year={2021}
}

@article{smith1957physical,
  title={Physical evidence in the investigation of traffic accidents},
  author={Smith, H Ward},
  journal={J. Crim. L. Criminology \& Police Sci.},
  volume={48},
  pages={93},
  year={1957},
  publisher={HeinOnline}
}

@article{gadotti2024anonymization,
  title={Anonymization: The imperfect science of using data while preserving privacy},
  author={Gadotti, Andrea and Rocher, Luc and Houssiau, Florimond and Cre{\c{t}}u, Ana-Maria and De Montjoye, Yves-Alexandre},
  journal={Science advances},
  volume={10},
  number={29},
  pages={eadn7053},
  year={2024},
  publisher={American Association for the Advancement of Science}
}

@article{morehouse2024responsible,
  title={Responsible data sharing: Identifying and remedying possible re-identification of human participants.},
  author={Morehouse, Kirsten N and Kurdi, Benedek and Nosek, Brian A},
  journal={American Psychologist},
  year={2024},
  publisher={American Psychological Association}
}

@article{hossain2023data,
  title={Data mining approach to explore emergency vehicle crash patterns: A comparative study of crash severity in emergency and non-emergency response modes},
  author={Hossain, Md Mahmud and Zhou, Huaguo and Das, Subasish},
  journal={Accident Analysis \& Prevention},
  volume={191},
  pages={107217},
  year={2023},
  publisher={Elsevier}
}

@article{chen2025transforming,
  title={Transforming traffic accident investigations: a virtual-real-fusion framework for intelligent 3D traffic accident reconstruction},
  author={Chen, Yanzhan and Zhang, Qian and Yu, Fan},
  journal={Complex \& Intelligent Systems},
  volume={11},
  number={1},
  pages={76},
  year={2025},
  publisher={Springer}
}

@article{jiao2018virtual,
  title={A virtual reality method for digitally reconstructing traffic accidents from videos or still images},
  author={Jiao, Peifeng and Miao, Qifeng and Zhang, Meichao and Zhao, Weidong},
  journal={Forensic science international},
  volume={292},
  pages={176--180},
  year={2018},
  publisher={Elsevier}
}

@article{font2012reconstruccion,
  title={Reconstrucci{\'o}n de accidentes:“la importancia del atestado oficial”},
  author={Font Mezquita, Jos{\'e}},
  journal={Securitas Vialis},
  volume={4},
  number={1},
  pages={9--15},
  year={2012},
  publisher={Springer}
}

@book{rider2017impact,
  title={The impact of new technology on crash reconstruction},
  author={Rider, Roger Ryan},
  year={2017},
  publisher={Tarleton State University}
}

@article{ball2005working,
  title={Working with images in daily life and police practice: an assessment of the documentary tradition},
  author={Ball, Mike},
  journal={Qualitative Research},
  volume={5},
  number={4},
  pages={499--521},
  year={2005},
  publisher={Sage Publications Sage CA: Thousand Oaks, CA}
}

@article{komter2006talk,
  title={From talk to text: The interactional construction of a police record},
  author={Komter, Martha L},
  journal={Research on Language and Social interaction},
  volume={39},
  number={3},
  pages={201--228},
  year={2006},
  publisher={Taylor \& Francis}
}

@article{hu2021research,
  title={Research on risky driving behavior evaluation model based on CIDAS real data},
  author={Hu, Lin and Bao, Xingqian and Lin, Miao and Yu, Chao and Wang, Fang},
  journal={Proceedings of the Institution of Mechanical Engineers, Part D: Journal of Automobile Engineering},
  volume={235},
  number={8},
  pages={2176--2187},
  year={2021},
  publisher={SAGE Publications Sage UK: London, England}
}

@article{otte2012injury,
  title={Injury protection and accident causation parameters for vulnerable road users based on German In-Depth Accident Study GIDAS},
  author={Otte, Dietmar and J{\"a}nsch, Michael and Haasper, Carl},
  journal={Accident Analysis \& Prevention},
  volume={44},
  number={1},
  pages={149--153},
  year={2012},
  publisher={Elsevier}
}

@inproceedings{kreiss2015extrapolation,
  title={Extrapolation of GIDAS accident data to Europe},
  author={Kreiss, Jens-Peter and Feng, Gang and Krampe, Jonas and Meyer, Marco and Niebuhr, Tobias and Pastor, Claus and Dobberstein, Jan},
  booktitle={Proceedings of the 24th ESV Conference Proceedings},
  year={2015},
  organization={NHTSA Washington, DC}
}

@book{tiwari2018transport,
  title={Transport planning and traffic safety: making cities, roads, and vehicles safer},
  author={Tiwari, Geetam and Mohan, Dinesh and Agrawal, Girish},
  year={2018},
  publisher={CRC Press}
}

@article{montella2013crash,
  title={Crash databases in Australasia, the European Union, and the United States: review and prospects for improvement},
  author={Montella, Alfonso and Andreassen, David and Tarko, Andrew P and Turner, Shane and Mauriello, Filomena and Imbriani, Lella Liana and Romero, Mario A},
  journal={Transportation research record},
  volume={2386},
  number={1},
  pages={128--136},
  year={2013},
  publisher={SAGE Publications Sage CA: Los Angeles, CA}
}

@article{zhang2018deep,
  title={A deep learning approach for detecting traffic accidents from social media data},
  author={Zhang, Zhenhua and He, Qing and Gao, Jing and Ni, Ming},
  journal={Transportation research part C: emerging technologies},
  volume={86},
  pages={580--596},
  year={2018},
  publisher={Elsevier}
}

@inproceedings{shin2024recap,
  title={RECAP: 3D traffic reconstruction},
  author={Shin, Christina Suyong and Pang, Weiwu and Li, Chuan and Bai, Fan and Ahmad, Fawad and Paek, Jeongyeup and Govindan, Ramesh},
  booktitle={Proceedings of the 30th Annual International Conference on Mobile Computing and Networking},
  pages={1252--1267},
  year={2024}
}

@article{van2020review,
  title={A review on the long short-term memory model},
  author={Van Houdt, Greg and Mosquera, Carlos and N{\'a}poles, Gonzalo},
  journal={Artificial intelligence review},
  volume={53},
  number={8},
  pages={5929--5955},
  year={2020},
  publisher={Springer}
}

@inproceedings{han2020stgcn,
  title={STGCN: a spatial-temporal aware graph learning method for POI recommendation},
  author={Han, Haoyu and Zhang, Mengdi and Hou, Min and Zhang, Fuzheng and Wang, Zhongyuan and Chen, Enhong and Wang, Hongwei and Ma, Jianhui and Liu, Qi},
  booktitle={2020 IEEE International Conference on Data Mining (ICDM)},
  pages={1052--1057},
  year={2020},
  organization={IEEE}
}

@article{nayakanti2022wayformer,
  title={Wayformer: Motion forecasting via simple \& efficient attention networks},
  author={Nayakanti, Nigamaa and Al-Rfou, Rami and Zhou, Aurick and Goel, Kratarth and Refaat, Khaled S and Sapp, Benjamin},
  journal={arXiv preprint arXiv:2207.05844},
  year={2022}
}

@book{schumaker2007spline,
  title={Spline functions: basic theory},
  author={Schumaker, Larry},
  year={2007},
  publisher={Cambridge university press}
}

@inproceedings{zhou2022hivt,
  title={Hivt: Hierarchical vector transformer for multi-agent motion prediction},
  author={Zhou, Zikang and Ye, Luyao and Wang, Jianping and Wu, Kui and Lu, Kejie},
  booktitle={Proceedings of the IEEE/CVF conference on computer vision and pattern recognition},
  pages={8823--8833},
  year={2022}
}

@inproceedings{gao2020vectornet,
  title={Vectornet: Encoding hd maps and agent dynamics from vectorized representation},
  author={Gao, Jiyang and Sun, Chen and Zhao, Hang and Shen, Yi and Anguelov, Dragomir and Li, Congcong and Schmid, Cordelia},
  booktitle={Proceedings of the IEEE/CVF conference on computer vision and pattern recognition},
  pages={11525--11533},
  year={2020}
}

@article{ali2025world,
  title={World simulation with video foundation models for physical ai},
  author={Ali, Arslan and Bai, Junjie and Bala, Maciej and Balaji, Yogesh and Blakeman, Aaron and Cai, Tiffany and Cao, Jiaxin and Cao, Tianshi and Cha, Elizabeth and Chao, Yu-Wei and others},
  journal={arXiv preprint arXiv:2511.00062},
  year={2025}
}

@article{krajzewicz2012recent,
  title={Recent development and applications of SUMO-Simulation of Urban MObility},
  author={Krajzewicz, Daniel and Erdmann, Jakob and Behrisch, Michael and Bieker, Laura and others},
  journal={International journal on advances in systems and measurements},
  volume={5},
  number={3\&4},
  pages={128--138},
  year={2012}
}

@inproceedings{zhang2006virtual,
  title={Virtual reconstruction of two types of traffic accident by the tire marks},
  author={Zhang, Xiaoyun and Jin, Xianlong and Shen, Jie},
  booktitle={International Conference on Artificial Reality and Telexistence},
  pages={1128--1135},
  year={2006},
  organization={Springer}
}

@techreport{steffan1996collision,
  title={The collision and trajectory models of PC-CRASH},
  author={Steffan, Hermann and Moser, Andreas},
  year={1996},
  institution={SAE Technical Paper}
}
\end{document}